\documentclass[manuscript]{acmart}

\setcopyright{rightsretained}
\copyrightyear{2025}
\acmYear{2025}

\usepackage{algorithmic}
\usepackage{graphicx}
\usepackage{textcomp}
\usepackage{xcolor}
\usepackage{siunitx}
\usepackage{makecell}
\usepackage{threeparttable}
\usepackage{enumitem}
\usepackage{booktabs}
\usepackage{hyperref}
\usepackage{multicol}
\usepackage{multirow}
\usepackage{tabularx}
\usepackage[toc,page]{appendix}

\PassOptionsToPackage{hyphens}{url}\usepackage{hyperref}

\newcolumntype{P}[1]{>{\centering\arraybackslash}p{#1}}

\title{wa-hls4ml: A Benchmark and Surrogate Models for hls4ml Resource and Latency Estimation}
\thanks{FERMILAB-PUB-25-0359-CSAID}

\author{Benjamin Hawks}
\orcid{0000-0001-5700-0288}
\affiliation{%
  \institution{Fermi National Accelerator Laboratory}
  \city{Batavia}
  \state{IL}
  \country{USA}}
\email{bhawks@fnal.gov}

\author{Jason Weitz}
\orcid{0009-0004-6315-3562}
\affiliation{%
  \institution{University of California San Diego}
  \city{La Jolla}
  \state{CA}
  \country{USA}}
\email{jdweitz@ucsd.edu}

\author{Dmitri Demler}
\orcid{0009-0009-9453-9755}
\affiliation{%
  \institution{University of California San Diego}
  \city{La Jolla}
  \state{CA}
  \country{USA}}
\email{ddemler@ucsd.edu}

\author{Karla Tame-Narvaez}
\orcid{0000-0002-2249-9450}
\affiliation{%
  \institution{Fermi National Accelerator Laboratory}
  \city{Batavia}
  \state{IL}
  \country{USA}}
\email{karla@fnal.gov}

\author{Dennis Plotnikov}
\orcid{0000-0002-2610-8226}
\affiliation{%
  \institution{Johns Hopkins University}
  \city{Baltimore}
  \state{MD}
  \country{USA}}
\email{dennis.yuri.plotnikov@cern.ch}

\author{Mohammad Mehdi Rahimifar}
\orcid{0000-0002-6582-8322}
\affiliation{%
  \institution{University of Sherbrooke}
  \city{Sherbrooke}
  \state{Quebec}
  \country{Canada}}
\email{rahm2701@usherbrooke.ca}

\author{Hamza Ezzaoui Rahali}
\orcid{0000-0002-0352-725X}
\affiliation{%
  \institution{University of Sherbrooke}
  \city{Sherbrooke}
  \state{Quebec}
  \country{Canada}}
\email{hamza.rahali@usherbrooke.ca}

\author{Audrey C. Therrien}
\orcid{0000-0001-6698-8400}
\affiliation{%
  \institution{University of Sherbrooke}
  \city{Sherbrooke}
  \state{Quebec}
  \country{Canada}}
\email{audrey.corbeil.therrien@usherbrooke.ca}

\author{Donovan Sproule}
\orcid{0009-0008-6719-5769}
\affiliation{%
  \institution{Columbia University}
  \city{New York}
  \state{NY}
  \country{USA}}
\email{donovan.sproule@gmail.com}

\author{Elham E Khoda}
\orcid{0000-0001-8720-6615}
\affiliation{%
  \institution{University of California San Diego}
  \city{La Jolla}
  \state{CA}
  \country{USA}}
\email{ekhoda@ucsd.edu}

\author{Keegan A. Smith}
\orcid{0009-0004-0653-7033}
\affiliation{%
  \institution{Texas A\&M University}
  \city{College Station}
  \state{TX}
  \country{USA}}
\email{keeganasmith2003@tamu.edu}

\author{Russell Marroquin}
\orcid{0000-0002-3364-7463}
\affiliation{%
  \institution{University of California San Diego}
  \city{La Jolla}
  \state{CA}
  \country{USA}}
\email{rdmarroquinsolares@ucsd.edu}

\author{Giuseppe Di Guglielmo}
\orcid{0000-0002-5749-1432}
\affiliation{%
  \institution{Fermi National Accelerator Laboratory}
  \city{Batavia}
  \state{IL}
  \country{USA}}
\email{gdg@fnal.gov}

\author{Nhan Tran}
\orcid{0000-0002-8440-6854}
\affiliation{%
  \institution{Fermi National Accelerator Laboratory}
  \city{Batavia}
  \state{IL}
  \country{USA}}
\email{ntran@fnal.gov}

\author{Javier Duarte}
\orcid{0000-0002-5076-7096}
\affiliation{%
  \institution{University of California San Diego}
  \city{La Jolla}
  \state{CA}
  \country{USA}}
\email{jduarte@ucsd.edu}

\author{Vladimir Loncar}
\orcid{0000-0003-3651-0232}
\affiliation{%
  \institution{European Organization for Nuclear Research (CERN)}
  \city{Geneva}
  \country{Switzerland}}
\email{vladimir.loncar@cern.ch}

\begin{abstract}
As machine learning (ML) is increasingly implemented in hardware to address real-time challenges in scientific applications, the development of advanced toolchains has significantly reduced the time required to iterate on various designs.
These advancements have solved major obstacles, but also exposed new challenges.
For example, processes that were not previously considered bottlenecks, such as hardware synthesis, are becoming limiting factors in the rapid iteration of designs.
To mitigate these emerging constraints, multiple efforts have been undertaken to develop an ML-based surrogate model that estimates resource usage of ML accelerator architectures.
We introduce wa-hls4ml, a benchmark for ML accelerator resource and latency estimation, and its corresponding initial dataset of over 680\,000 fully connected and convolutional neural networks, all synthesized using hls4ml and targeting Xilinx FPGAs.
The benchmark evaluates the performance of resource and latency predictors against several common ML model architectures, primarily originating from scientific domains, as exemplar models, and the average performance across a subset of the dataset.
Additionally, we introduce GNN- and transformer-based surrogate models that predict latency and resources for ML accelerators.
We present the architecture and performance of the models and find that the models generally predict latency and resources for the 75\% percentile within several percent of the synthesized resources on the synthetic test dataset.
\end{abstract}



\begin{document}

\maketitle

\section{Introduction}
Domain-specific design tools leveraging high-level synthesis (HLS) have become indispensable in the design of hardware accelerators, enabling the automatic translation of high-level programming languages like C++ or Python into hardware descriptions.
This shift allows developers to focus on algorithmic codesign rather than the complexities of hardware implementation, reducing both development time and the expertise required.
However, one of the major challenges in using HLS tools is predicting resource utilization---such as logic elements, memory, and interconnects---during the codesign process.
This requires costly and time-extensive hardware synthesis steps---both at the C and logic synthesis steps.
C synthesis is required for timing estimates, its resource estimation is typically inaccurate.
While logic synthesis produces accurate resource estimates, it takes a considerable amount of time.

Specialized domains like machine learning (ML), where hardware efficiency is paramount, magnify this challenge, as synthesis takes significant time especially for complex ML models.
One proposed solution to greatly accelerate the resource estimation step is a neural network surrogate model predictor, which can take the resource prediction step from hours to seconds.
A surrogate model is a model built to approximate a larger, more complex system.
In this case it is a neural network designed to approximate the resource estimation.

Developing generalized and high-performance surrogate models is challenging, both in terms of dataset generation and model design.
To address this at a more focused, initial scale, we consider dataflow FPGA architectures for embedded AI applications such as the internet of things (IoT), autonomous vehicles, and scientific sensing.
Within this scope, we aim to develop a surrogate model using the hls4ml flow~\cite{duarte2018fast}.
This allows designers to make better-informed decisions early in the development cycle, on the order of seconds from model specification, reducing the need for iterative synthesis runs and enabling more efficient hardware implementations.
For hls4ml users, this is particularly beneficial, as it provides detailed feedback on the hardware requirements of neural network architectures, helping developers optimize their models for FPGA deployment.

hls4ml is an open-source framework that translates ML models into FPGA-based IP using HLS tools.
It bridges the gap between ML and hardware by facilitating the development of low-latency, resource-efficient inference engines on FPGAs.
Despite these advantages, optimizing resource usage remains a complex task, requiring accurate predictions of hardware demands to effectively balance performance, power, and area constraints.

\begin{figure}
    \centering
    \includegraphics[width=0.9\linewidth,page=1]{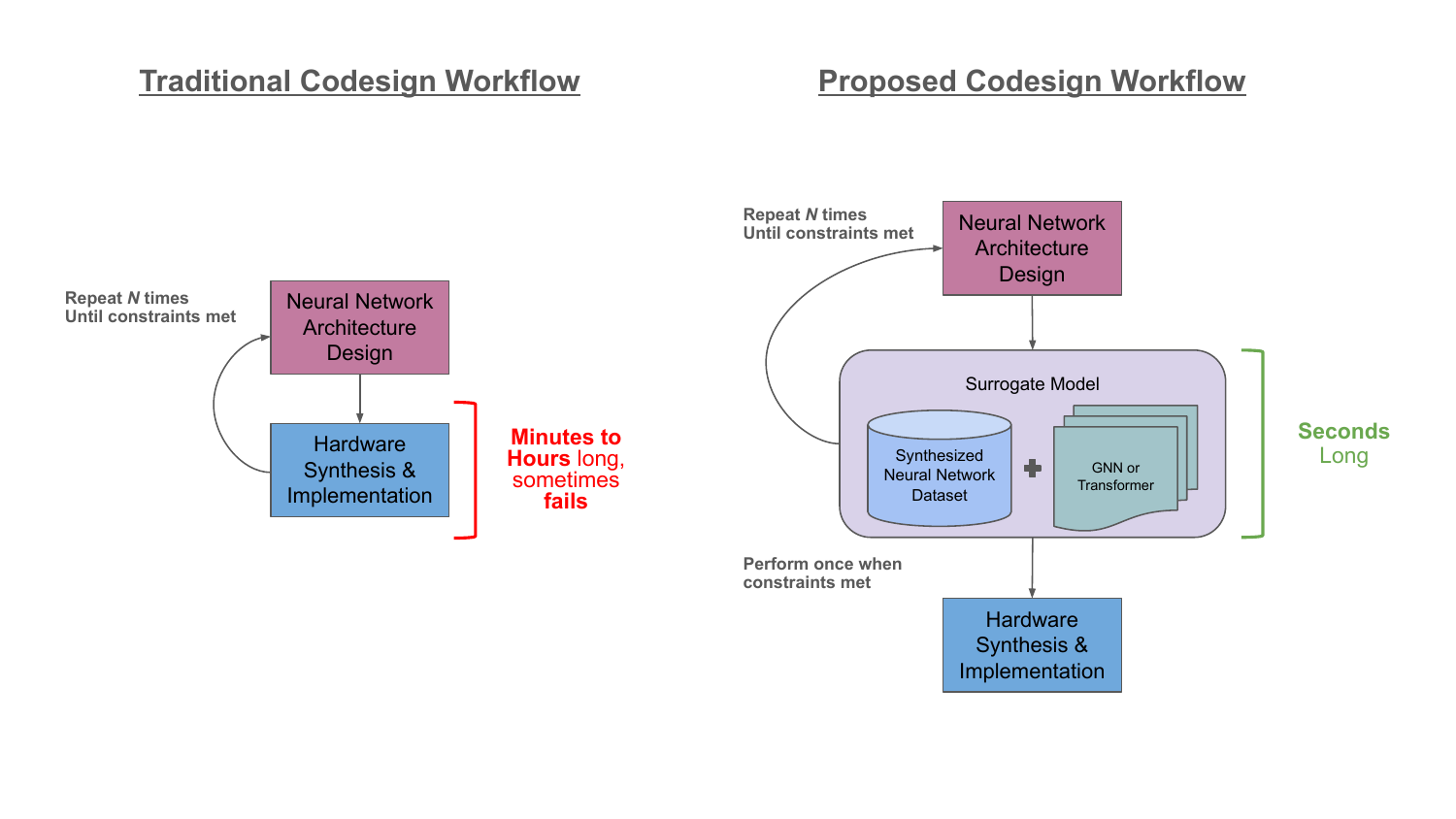}
    \caption{The traditional codesign workflow compared to the proposed surrogate model based codesign workflow.}
    \label{fig:overview-fig}
\end{figure}

To achieve this, we introduce \textbf{wa-hls4ml}\footnote{Named after Wario and Waluigi who are doppelg\"{a}ngers of Mario and Luigi, respectively, in the Nintendo Super Mario platform game series.}: a \textbf{dataset} unprecedented in scale and features, 
a \textbf{benchmark} for common evaluation,
and two new \textbf{surrogate models} for flexible and precise resource and latency estimation targeting the prediction of resource usage and latency for HLS tools. 
The combination of these three main contributions enables novel rapid codesign research at a scale beyond previously possible, by reducing the codesign loop from hours to seconds as shown in \autoref{fig:overview-fig}, and is a unique community resource.
Furthermore, it enables users of hls4ml and other dataflow accelerators for edge ML applications to rapidly deploy optimal FPGA implementations.

The open \textbf{dataset} is unprecedented in terms of its size, with over 680\,000 fully synthesized dataflow models.
The goal is to continue to grow and extend the dataset over time.
We include all steps of the synthesis chain from ML model to HLS representation to register-transfer level (RTL) and save the full logs.
This will enable a much broader set of applications beyond those in this paper.  
The \textbf{benchmark} standardizes evaluation of the performance of resource usage and latency estimators across a suite of metrics, such as the coefficient of determination ($R^2$), symmetric mean absolute percentage error (SMAPE), and root mean square error (RMSE), and provides sample models, both synthetic and from scientific applications, to support and encourage the continued development of better surrogate models.
The \textbf{surrogate model} architectures are based on a graph NN and transformer to enable flexibility.  The models predict FPGA resources: lookup tables (LUTs), flip-flops (FFs), digital signal processors (DSPs), and on-chip block random access memory (BRAM), as well as latency (clock cycles) and initiation interval (II).

The following summarizes our design rationale when it comes to building and maintaining the \textbf{benchmark}:
\begin{itemize}[noitemsep, topsep=0pt]
    \item \textbf{Address the need for a standard evaluation suite}: Provide exemplar benchmark models with their synthesis results, artifacts, and log files, along with comprehensive and predefined evaluation metrics.
    \item \textbf{Formalize the structure for optimal utility}: The benchmark will be structured to maximize its utility for the broader research community, ensuring that it remains applicable beyond resource and latency estimation.
    \item \textbf{Promote an open design process}: Allow contributors to propose enhancements to the benchmark and submit new surrogate models, in alignment with the latest advancements in the ML field.
\end{itemize}
In this way, we plan for the dataset and procedures laid out in this work to be a community benchmark for future avenues of study.  

\subsection{Related Work}
Previous HLS design datasets have focused on more generic lower-level kernels, such as general matrix multiplication.
For example, DB4HLS~\cite{db4hls} targets programs from the MachSuite benchmark~\cite{machsuite}, GNN-DSE~\cite{gnn-dse} targets programs from
PolyBench/C~\cite{polyhedral}, while \textsc{HLSyn}~\cite{hlsyn} includes kernels from both. 
In contrast, the wa-hls4ml dataset is more domain-specific, though it also targets a large variety of multilayer neural network designs generated by hls4ml, and thus incorporates higher-level programs.
Concurrently with this work, HLSFactory~\cite{hlsfactory} is a framework to collect and build HLS design datasets, including ML designs generated by FlowGNN~\cite{flowgnn}.
Finally, rule4ml~\cite{rahimifar2024rule4ml} is a closely related prior work from which we expand the dataset, formalize a benchmark, and explore more complex surrogate model architectures.
Other datasets and surrogate models exist~\cite{8457644}, \cite{9835440}, \cite{10213402} but are not open source or have relatively smaller dataset sizes. Vivado/Vitis HLS also provide native estimates~\cite{AMD2024VitisHLS}, but only after running C-synthesis, which can be time-consuming.
\autoref{tab:related_work} summarizes directly comparable studies and tools, laying out the datasets, the representation used as input, the availability of the code, and whether the approach proposes a benchmark.

\newcolumntype{H}[1]{>{\centering\arraybackslash}p{#1}}
\begin{table*}[t]
    \centering
    \caption{Comparison of wa-hls4ml to prior work. }
    \label{tab:related_work}
    \resizebox{\textwidth}{!}{%
    \begin{tabular}{cH{1cm}H{1cm}H{1.5cm}H{2cm}H{1.5cm}H{3.5cm}}
    \toprule
        \textbf{Tool} & \textbf{Open Source} & \textbf{Open Dataset} & \textbf{Dataset Size [samples]} & \textbf{Vendor tools required} & \textbf{Is Benchmark}  & \textbf{Input Abstraction Level}\\ 
    \midrule
        \makecell{Native HLS Estimate ~\cite{AMD2024VitisHLS}}  & No & N/A & N/A & Yes & N/A &  \makecell{HLS Code}\\
        
        \makecell{High-Level Synthesis Performance \\ Prediction using GNNs~\cite{10.1145/3489517.3530408}}& \textbf{Yes} & \textbf{Yes} & $40,000$ & Yes & \textbf{Yes} & \makecell{HLS/LLVM IR Graph} \\ 
        
        \makecell{Machine Learning Aided Hardware \\ Resource Estimation for FPGA \\ DNN Implementation~\cite{9835440}} & No & No & N/A & \textbf{No} & No & \makecell{FINN Intermediate \\ Representation} \\
        
        \makecell{A Graph Neural Network Model for Fast \\ and Accurate Quality of Result Estimation \\ for High-Level Synthesis~\cite{10213402}} & No & No & $2,465$ & Yes & No &  \makecell{HLS/LLVM IR Graph} \\
        
        \makecell{HLSyn ~\cite{hlsyn}}& \textbf{Yes} & \textbf{Yes} & $42,000$ &\textbf{No} & \textbf{Yes} & \makecell{HLS Code} \\ 
        
        \makecell{Fast and Accurate Estimation \\of Quality of Results in \\High-Level Synthesis \\with Machine Learning~\cite{8457644} } & No & No & $1,300$ & Yes & No  & \makecell{HLS Reports} \\
        
    \midrule
        \makecell{rule4ml~\cite{rahimifar2024rule4ml} (Related Work)} & \textbf{Yes} & \textbf{Yes} & \textbf{$15,000$} & \textbf{No} & No &  \makecell{hls4ml IR} \\
        \makecell{wa-hls4ml (This work)} & \textbf{Yes} & \textbf{Yes} & \textbf{$683,176$} & \textbf{No} & \textbf{Yes} &  \makecell{hls4ml IR} \\
        
    \bottomrule
    \end{tabular}%
    }
\end{table*}

Since we target higher-level hls4ml programs, it is difficult to make direct comparisons to our method.
Additionally, in many cases of related work, the code is not available for evaluation.
The work by Wu et al.~\cite{10.1145/3489517.3530408} has publicly available code and most closely aligns with our approach in terms of predicting FPGA resource utilization and timing from high-level descriptions using graph neural networks (GNNs). Their framework processes intermediate representation (IR) graphs extracted after front-end compilation of C/C++ programs, enabling resource usage and timing estimation without completing the entire HLS process.

While conceptually similar, our approaches differ significantly in several key ways.
First,~\cite{10.1145/3489517.3530408} targets general C/C++ applications synthesized with Vitis HLS, with their training dataset comprising programs generated by C code generator ldrgen~\cite{barany2017liveness}, and applications from PolyBench/C, CHStone~\cite{hara2009proposal}, and MachSuite~\cite{reagen2014machsuite}.
Conversely, our work specifically focuses on neural networks implemented using hls4ml.
Second, their model necessitates running the initial stages of C synthesis to extract IR graphs, whereas our GNN- and transformer-based surrogate models operate directly on the neural network architecture description.
This eliminates synthesis steps, providing faster resource prediction.

Because the wa-hls4ml dataset includes the HLS LLVM IR graphs, we can evaluate the approach presented in~\cite{10.1145/3489517.3530408} using their publicly available code.
Since our evaluation dataset is outside of their training domain, the predictions are not directly interpretable, e.g., many of the predictions are negative.
However, they still show a correlation to the ground truth.
To quantitatively compare the approaches, we adapted their model to our task by applying linear correction factors to their predictions when evaluating neural network models for simple 2-layer MLPs.
Even after this correction, their predictions are less precise compared to our GNN- and transformer-based surrogate models.
Their model achieved SMAPE values of 34.30, 36.03, and 31.26\% for DSPs, LUTs, and FFs, respectively.
These results indicate that domain-specific approaches like ours deliver better accuracy for ML workloads compared to general-purpose HLS estimation frameworks while also bypassing synthesis, further reducing estimation time.
An interesting future study would be to train their model on the wa-hls4ml dataset and investigate whether their approach improves prediction accuracy.  

\section{Dataset}
The dataset has two primary components, each designed to test different aspects of a surrogate model's performance.
The first part is based on \textit{synthetic} neural networks generated with various layer types, micro-architectures, and precisions.
This synthetic dataset lets us systematically explore the FPGA resources and latencies as we vary different model parameters.
The second part of the benchmark targets models from \textit{exemplar realistic} scientific applications, requiring real-time processing at the edge, near the data sources.
Models with real-time constraints constitute a primary use case for ML-to-FPGA pipelines like hls4ml.
This part tests the ability of the surrogate model to extrapolate its predictions to new configurations and architectures beyond the training set, assessing the model's robustness and performance for real applications.

The training, validation, and test sets of the benchmark currently consist of 683\,176 synthetic samples, consisting of data about synthesized samples of 608\,679 fully-connected neural networks, 31\,278 one-dimensional convolutional neural networks, and 43\,219  two-dimensional convolutional neural networks.
Each sample contains the model architecture, hls4ml~\cite{duarte2018fast} conversion parameters, the latency and resource usage numbers for that network post-logic synthesis, and associated metadata.
In addition to the training, validation, and test sets, the dataset also includes 887 samples representing the successful logic synthesis of the exemplar models with varying hls4ml conversion parameters, as shown in \autoref{subsubsec:datagen}.
The dataset as a whole is split, distributed, and intended to be used as follows:
\begin{itemize}
    \item \textbf{Training set}: The set of 478\,220 samples intended to be used for training a given estimator.
    \item \textbf{Validation set}: The set of 102\,472  samples intended to be used for validation during training.
    \item \textbf{Test set}: The set of 102\,484 samples intended to be used for testing and generating results for a given estimator.
    \item \textbf{Exemplar test set}: The set of 887 samples, comprising the models described in \autoref{subsec:benchmodels}, intended to be used for testing and generating results for a given estimator.
\end{itemize}
Within each subset, excluding the exemplar test set, the data is further grouped as follows.
These categories explain the composition of our dataset but have no bearing on how a given estimator should be trained.
\begin{itemize}
    \item \textbf{2\_20}: The updated rule4ml dataset, containing fully-connected neural networks that were randomly generated with layer counts between 2 and 20 layers, using hls4ml resource and latency strategies. 
    \item \textbf{2\_layer}: a subset containing 2-layer deep fully-connected neural networks generated via a grid search using hls4ml resource and io\_parallel strategies 
    \item \textbf{3\_layer}: a subset containing 3-layer deep fully-connected neural networks generated via a grid search using hls4ml resource and io\_parallel strategies 
    \item \textbf{conv1d}: A subset containing 3--7 layer deep 1-dimensional convolutional neural networks that were randomly generated and use hls4ml resource and io\_stream strategies 
    \item \textbf{conv2d}: A subset containing 3--7 layer deep 2-dimensional convolutional neural networks that were randomly generated and use hls4ml resource and io\_stream strategies 
    \item \textbf{latency}: a subset containing 3--7 layer deep fully-connected neural networks that were randomly generated and use hls4ml latency and io\_parallel strategies 
    \item \textbf{resource}: a subset containing 3--7 layer deep fully-connected neural networks that were randomly generated and use hls4ml resource and io\_parallel strategies 
\end{itemize}

\subsection{Synthetic Dataset}
\label{subsec:dataset}

With the introduction of ML into FPGA toolchains, e.g. for resource and latency prediction or code generation, there is a significant need for large datasets to support and train these tools.
We found that existing datasets were insufficient for these needs, and therefore sought to build a dataset and a highly scalable data generation framework that is useful for a wide variety of research surrounding ML on FPGAs.
This dataset serves as one of the few openly accessible, large-scale collections of synthesized neural networks available for ML research. 

\subsubsection{Generation}
\label{subsubsec:datagen}
The train and test sets were created by first generating models of varying architectures in the Keras and QKeras~\cite{qkeras} Python libraries, varying their hyperparameters.
The updated rule4ml dataset follows the same generation method and hyperparameter ranges described in~\cite{rahimifar2024rule4ml}, while adding II information and logic synthesis results to the reports.

For the remaining subsets of the data, the two-layer and three-layer fully-connected models were generated using a grid search method according to the parameter ranges mentioned below, 
whereas larger fully-connected models and convolutional models (one- and two-dimensional) were randomly generated
, where convolutional modes also contain dense, flatten, and pooling layers.
The weight and bias precision was implemented in HLS as datatype \texttt{ap\_fixed<$X$,1>}, where $X$ is the specified precision and the total number of bits allocated to the weight and bias values, with one bit being reserved for the integer portion of the value.
These models were then converted to HLS using hls4ml and synthesized through AMD Vitis version 2023.2 and 2024.2, targeting the AMD Xilinx Alveo U250 FPGA board~\cite{AlveoU250datasheet}. 
The model sets have the following parameter ranges: 
\begin{itemize}[noitemsep, topsep=1pt]
    \item \textit{Number of layers}: 2-7 for fully-connected models; 3--7 for convolutional models
    \item \textit{Activation functions}: linear for most 2-3 layer fully-connected models; ReLU, $\tanh$, and sigmoid for all other fully connected models and convolutional models
    \item \textit{Number of features/neurons}: 8--128 (step size: 8 for 2--3 layer) for fully-connected models; 32--128 for convolution models with 8--64 filters
    \item \textit{Weight and bias bit precision}: 2--16 bits (step size: 2) for 2-3 layer fully-connected models, 4--16 bits (step size: powers of 2) for 3--7 layer fully-connected and convolutional models
    \item \textit{hls4ml target reuse factor}: 1--4093 for fully-connected models ; 8192--32795 for convolutional models
    \item \textit{hls4ml implementation strategy}: Resource strategy, which controls the degree of parallelism by explicitly specifying the number of MAC operations performed in parallel per clock cycle, is used for most fully-connected models and all convolutional models, while Latency strategy, where the computation is unrolled, is used for some 3--7 layer fully-connected models. 
    \item \textit{hls4ml I/O type}: The io\_parallel setting, which directly wires the output of one layer to the input of the next layer, is used for all fully-connected models, and the io\_stream setting, which uses FIFO buffers between layers, is used for all convolutional models.
\end{itemize}

The synthesis was repeated multiple times, varying the hls4ml \emph{reuse factor}, a tunable setting that proportionally limits the number of multiplication operations used.
The hls4ml conversion, HLS synthesis, and logic synthesis of the train and test sets were all performed in parallel on the National Research Platform Kubernetes Hypercluster and the Texas A\&M ACES HPRC Cluster.
On the National Research Platform, synthesis was run inside a container with a guest OS of Ubuntu 20.04.4 LTS, the containers being slightly modified versions of the xilinx-docker~\cite{xilinx-docker} v2023.2 ``user" images, with 3 virtual CPU Cores and 16~GB of RAM per pod, with all AMD tools mounted through a Ceph~\cite{Ceph}-based persistent volume.
Jobs run on the Texas A\&M ACES HPRC Cluster were run using Vitis 2024.2, each with 2 virtual CPU cores and 32~GB of RAM. 
The resulting projects, reports, logs, and a JSON file containing the resource/latency usage and estimates of the C and logic synthesis were collected for each sample in the dataset.
The data pertaining to the resource utilization and latency, neural network architecture, and information relating to the conversion using hls4ml was then further processed into a collection of JSON files, distributed alongside this paper and described below. 
The full projects, which contain the generated code, logs, intermediate representations, and other related files, are also available for each sample in the primary dataset as a part of a companion dataset also released alongside this paper. 

\subsubsection{Dataset Analysis}

\begin{figure}
    \centering
    \includegraphics[width=0.8\linewidth]{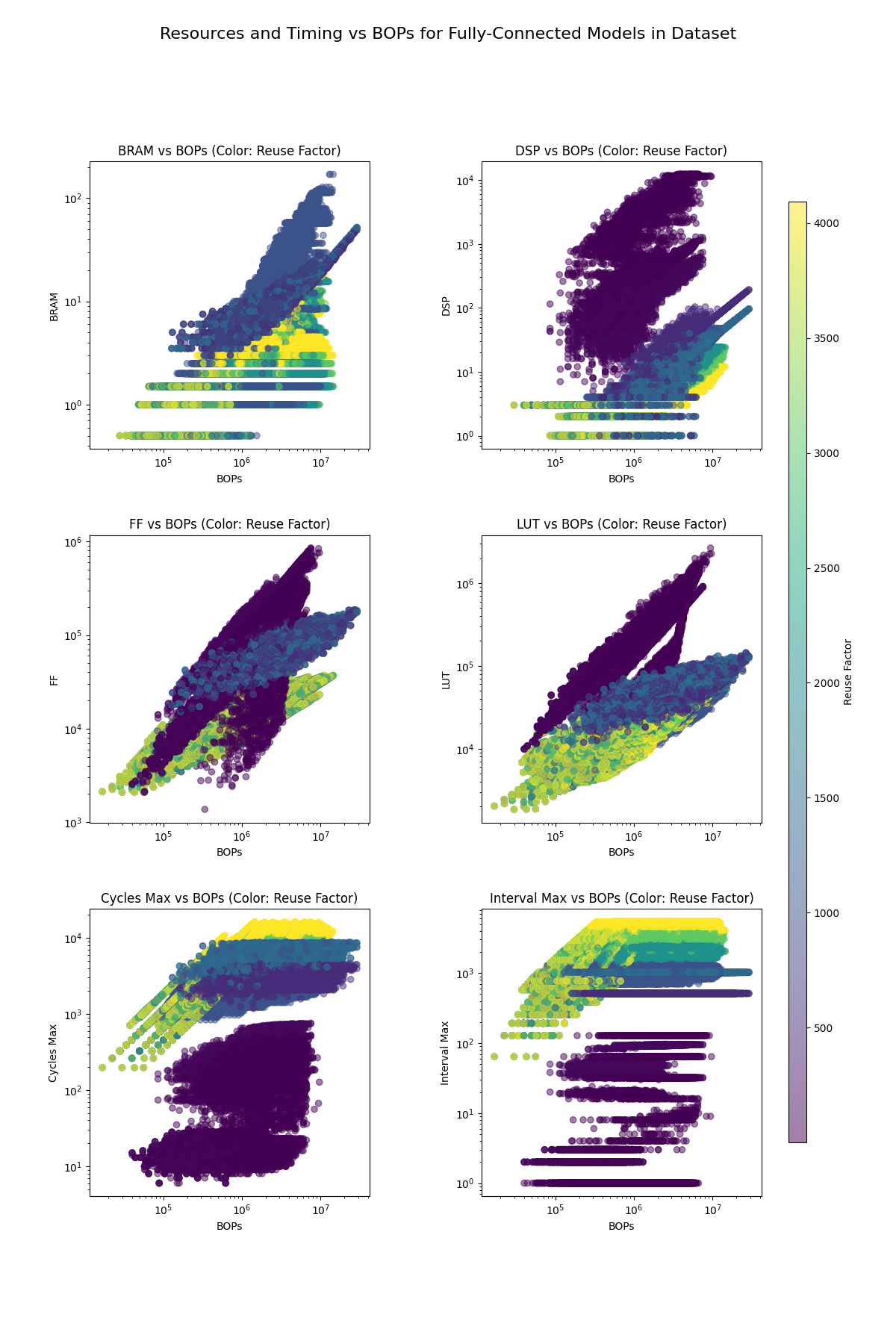}
    \caption{All tracked output features plotted for each fully-connected model in the dataset versus Bit-Operations, with the color representing the reuse factor. }
    \label{fig:dense-dataset}
\end{figure}

\begin{figure}
    \centering
    \includegraphics[width=0.8\linewidth]{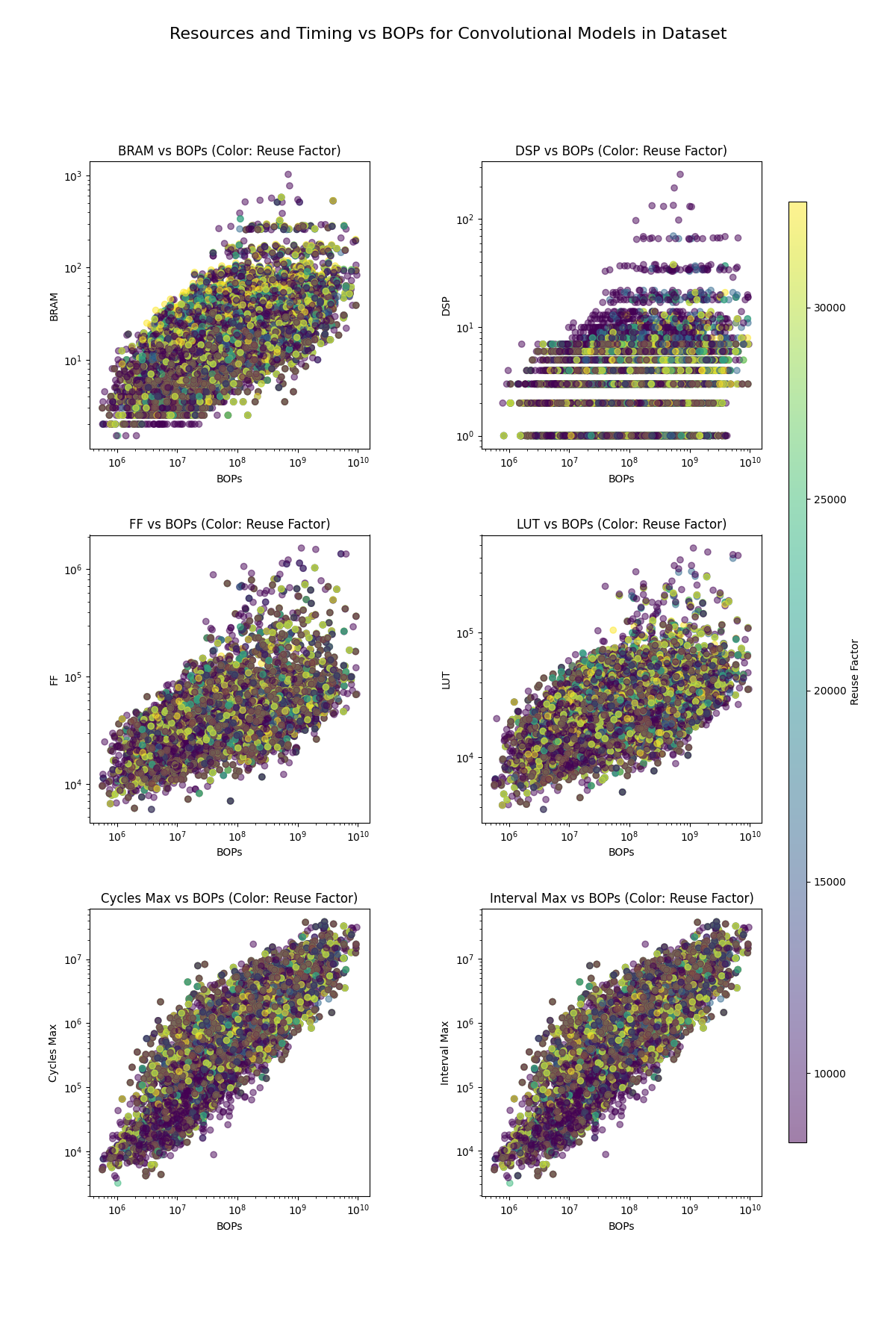}
    \caption{All tracked output features plotted for each convolutional model in the dataset versus Bit-Operations, with the color representing the reuse factor of a given sample}
    \label{fig:conv-dataset}
\end{figure}

We performed an analysis of the primary dataset to potentially identify trends and compare commonly used resource estimation metrics against actual resource utilization.
Below are selected figures, \autoref{fig:dense-dataset} and \autoref{fig:conv-dataset}, visualizing the dataset's resource and latency values against bit operations (BOPs)~\cite{BOP_Javi} and the reuse factor of a given model.

We observe that within the dataset, the BOPs metric tends to approximate the resource and timing values of fully-connected models more closely than convolutional models.
Within the convolutional models of the dataset, there is some positive correlation with BOPs, with the correlation being slightly better for timing estimates than resource information, but overall, the correlation is not as strong as it is for fully-connected models. 

Additionally, we note that fully-connected models have distinct populations when grouped according to the reuse factor.
These populations tend to be highly correlated for timing and resource information, with higher reuse factors exhibiting the expected behavior of larger latencies and lower resource usage in most cases.
This trend is not as strong for convolutional models, where the reuse factor is less impactful than model size and complexity when implementing the network on an FPGA.

\begin{figure}
    \centering
    \includegraphics[width=0.9\linewidth]{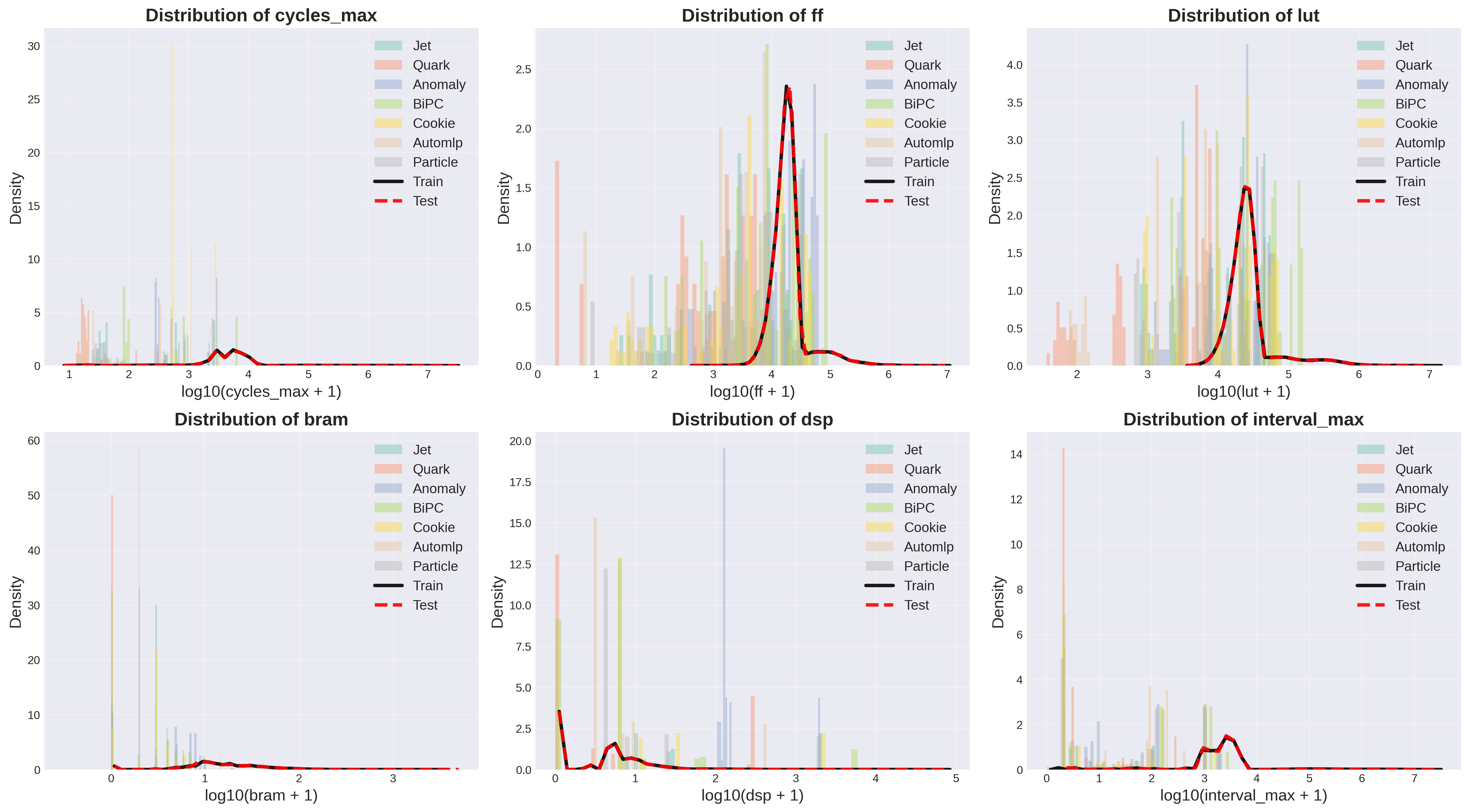}
    \caption{Exemplar vs the train and test subset resource (label) distributions}
    \label{fig:exemplar-label-distribution}
\end{figure}

We also visualize the distribution of resource and latency features throughout the test dataset and exemplar dataset  in~\autoref{fig:exemplar-label-distribution}. We find that the distributions between the exemplar models and test dataset tend to not overlap strongly, indicating that there is room to improve both the test dataset and exemplar dataset in terms of model architecture diversity, which is an area that we aim to improve upon in future works.


\subsubsection{Dataset Structure}
The distributed JSON files contain 683\,176 total samples. The samples are split into three subsets, as described in \autoref{subsec:dataset}. The format across the three subsets is the same, where each sample is a single JSON file containing 9 fields:

\begin{itemize}
    \item \texttt{meta\_data}: a unique identifier, model name, and name of the corresponding gzipped tarball of the fully synthesized project, logs, and reports for the sample (contained in an accompanying dataset released alongside the primary dataset)
    \item \texttt{model\_config}: a JSON representation of the Keras/QKeras~\cite{qkeras} model synthesized in the sample, including the actual reuse factor as synthesized per layer.
    \item \texttt{hls\_config}: the hls4ml configuration dictionary used to convert the model for the sample, including the target reuse factor as synthesized per layer
    \item \texttt{resource\_report}: a report of the post-logic synthesis resources used for the sample, reported as the actual number of components used.
    \item \texttt{hls\_resource\_report}: a report of the post-hls synthesis resources estimated for the sample, reported as the actual number of components estimated.
    \item \texttt{latency\_report}: a report of the post-hls synthesis latency estimates for the sample.
    \item \texttt{target\_part}: the FPGA part targeted for HLS and logic synthesis for the sample.
    \item \texttt{vivado\_version}: the version of Vivado used to synthesize the sample.
    \item \texttt{hls4ml\_version}: the version of hls4ml used to convert the sample.
\end{itemize}

\subsection{Exemplar Realistic Models}
\label{subsec:benchmodels}

The exemplar models utilized in this study include several key architectures, each tailored for specific ML tasks and targeting scientific applications with low-latency constraints.
The synthesis parameters for these models are presented in \autoref{tab:exemplar-params}.

\begin{table}
\renewcommand{\arraystretch}{1.6}
\centering
\caption{The hyperparameters used in the synthesis of the exemplar benchmark models.}
\label{tab:exemplar-params}
\begin{tabular}{c|P{0.9in}}
    \textbf{Hyperparameter} & \textbf{Values} \\ 
    \hline
    \textbf{Precision} & \texttt{ap\_fixed<2,1>}, \texttt{ap\_fixed<8,3>}, \texttt{ap\_fixed<16,6>} \\
    \textbf{Strategy} & \texttt{Latency}, \texttt{Resource} \\
    \textbf{Target reuse factor} & 1, 128, 1024 \\
    \textbf{Target board} & \ Alveo U200, \newline Alveo U250 \\
    \textbf{Target clock} & \SI{5}{ns}, \SI{10}{ns} \\
    \textbf{Vivado version} & 2019.1, 2020.1 \\
\end{tabular}
\end{table}

The following gives a brief description of each of these models and their applications, while \autoref{tab:exemplar-arch} presents their architectures. ``Jet''~\cite{hls4ml_named} is a fully connected neural network that classifies simulated particle jets originating from one of five particle classes in high-energy physics experiments. ``Quarks''~\cite{duarte2018fast} is a binary classifier for top quark jets. It helps probe fundamental particles and their interactions.
``Anomaly''~\cite{borras2022open} is an autoencoder trained on audio data to reproduce the input spectrogram, whose loss value differentiates between normal and abnormal signals. ``BiPC''~\cite{rahali2024efficient} refers to an encoder that transforms high-resolution images, producing sparse codes for further compression. ``Cookie''~\cite{gouin2022combining} is dedicated to real-time data acquisition for the CookieBox system, designed for advanced experimental setups requiring rapid handling of large data volumes generated by high-speed detectors.
``AutoMLP'' refers to a fully connected network from the AutoMLP framework~\cite{chen2023automlp}, which focuses on accelerating MLPs on FPGAs, providing significant improvements in computational performance and energy efficiency.
Lastly, ``Particle Tracking''~\cite{abidi2022charged} tracks charged particles in real-time as they traverse silicon detectors in large-scale particle physics experiments.

\begin{table}[t]
\small
\renewcommand{\arraystretch}{1.9}
\setlength\tabcolsep{3.65pt}
\centering
\caption{Architectures of the exemplar benchmark models.}
\label{tab:exemplar-arch}
\begin{tabular}{P{0.8in}|c|c|c}
Model & Size & Input & Architecture \\ \hline
Jet~\cite{duarte2018fast}
& 2,821 & 16 & $\xrightarrow[\text{ReLU}]{32}\ \xrightarrow[\text{ReLU}]{32}\ \xrightarrow[\text{ReLU}]{32}\ \xrightarrow[\text{Softmax}]{5}$ \\

Top Quarks~\cite{duarte2018fast}
& 385 & 10 & $\xrightarrow[\text{ReLU}]{32}\ \xrightarrow[\text{Sigmoid}]{1}$ \\

Anomaly~\cite{borras2022open}
& 2,864 & 128 & $\xrightarrow[\text{ReLU}]{8}\ \xrightarrow[\text{ReLU}]{4}\ \xrightarrow[\text{ReLU}]{128}\ \xrightarrow[\text{ReLU}]{4}\ \xrightarrow[\text{Softmax}]{128}$ \\

BiPC~\cite{rahali2024efficient}
& 7,776 & 36 & $\xrightarrow[\text{ReLU}]{36}\ \xrightarrow[\text{ReLU}]{36}\ \xrightarrow[\text{ReLU}]{36}\ \xrightarrow[\text{ReLU}]{36}\ \xrightarrow[\text{ReLU}]{36}$ \\

CookieBox~\cite{gouin2022combining}
& 3,433 & 512 & $\xrightarrow[\text{ReLU}]{4}\ \xrightarrow[\text{ReLU}]{32}\ \xrightarrow[\text{ReLU}]{32}\ \xrightarrow[\text{Softmax}]{5}$ \\

AutoMLP~\cite{chen2023automlp} 
& 534 & 7 & $\xrightarrow[\text{ReLU}]{12}\ \xrightarrow[\text{ReLU}]{16}\ \xrightarrow[\text{ReLU}]{12}\ \xrightarrow[\text{Softmax}]{2}$ \\

Particle Tracking~\cite{abidi2022charged} 
& 2,691 & 14 & $\xrightarrow[\text{ReLU}]{32}\ \xrightarrow[\text{ReLU}]{32}\ \xrightarrow[\text{ReLU}]{32}\ \xrightarrow[\text{Softmax}]{3}$ \\
\end{tabular}
\end{table}

\section{Benchmark}
\subsection{Submission Guidelines}
One important aspect in formalizing the benchmark structure is clearly defining the expected outputs, report format, and content guidelines for the contributors who plan to submit their surrogate models. This is vital to ensure consistent, reproducible, and fair evaluations. The following outlines the required, strongly recommended, and suggested components for a valid submission using the benchmark. The dataset and code availability for the benchmark are discussed in \autoref{sec:data_code}.

When submitting surrogate models for evaluation, contributors must provide a detailed report that includes the predicted values for each FPGA metric in the benchmark. The report should also include visual comparisons between predicted and actual values, presented in the form of box plots to demonstrate the model's accuracy. In addition, we strongly recommend that the report include a comprehensive description of the surrogate models' architectures, outlining key design details and hyperparameters used for training. While not required, sharing source code and trained weights is strongly encouraged to promote transparency and reproducibility of the results. We also recommend sharing the hardware specifications of the inference machine, along with the inference times. Lastly, we require documenting any further constraints, including additional training data (e.g., models, hls4ml configurations, or target boards), precision settings, or specific optimization strategies used during evaluation.

It is worth noting that we plan for the submission process to be open and ongoing, with no fixed release schedule. Instead, the benchmark will be updated to reflect significant contributions.

\subsection{Benchmark Metrics}
\label{subsec:benchmark_metrics}

After establishing the rationale and structure of the benchmark, we formalize the various metrics we apply to assess the performance of surrogate models on both the test set and the exemplar models. The metrics we use are as follows~\cite{chicco2021coefficient}:
\begin{itemize}
\item Coefficient of determination ($R^2$):
\begin{equation}
R^2 = 1 - \frac{\sum_{i=1}^{n} (y_i - \hat{y}_i)^2}{\sum_{i=1}^{n} (y_i - \bar{y})^2}
\end{equation}
\item Symmetric mean absolute percentage error (SMAPE):
\begin{equation}
\text{SMAPE} = \frac{200\%}{n} \sum_{i=1}^{n} \frac{|y_i - \hat{y}_i|}{|y_i| + |\hat{y}_i| + 1}
\end{equation}
\item Root mean square error (RMSE):
\begin{equation}
\text{RMSE} = \sqrt{\frac{1}{n} \sum_{i=1}^{n} (y_i - \hat{y}_i)^2} 
\end{equation}
\end{itemize}

In the above equations, $y_i$ represents the ground truth, $\hat{y}_i$ is the predicted value, and $\bar{y}$ is the mean of the ground truth values. The $R^2$ score evaluates the general performance of a predictor, measuring how well it captures the variability in the data. SMAPE offers insight into the relative accuracy of the predictions and is particularly useful when comparing errors across different scales. RMSE measures the magnitude of the prediction error, with sensitivity to larger outliers. In the case of SMAPE, we use the standard formula and add a small value $\epsilon$ to the denominator to avoid division by zero. We set $\epsilon$ to the smallest strictly positive value that the resource and latency variables can have, which in our case is 1.

For the evaluation of these metrics, the latency is measured in clock cycles, and resources are measured in absolute terms, rather than percentage utilization. Furthermore, in the current version of the benchmark, each performance metric is computed separately for each regression variable, ensuring a detailed evaluation across all prediction targets. In addition to these metrics, we use a box plot per variable to visualize the distribution of relative percentage errors (RPE), which we define as:
\begin{equation}
\text{RPE} = \left(\frac{y_i - \hat{y}_i}{y_i + 1} \right)\times 100\%
\end{equation}

\noindent The RPE box plots allow us to measure not only the spread of residuals but also identify trends in a surrogate model's predictions, providing a visual indication of whether a model tends to systematically underpredict or overpredict values.

\section{Synthesis surrogate model}
The development of the GNN and transformer surrogate models is another primary goal of this work. 
There is a significant challenge in designing an effective estimator that works for arbitrary neural network architectures.
Many different layer structures can be used, which may have radically different implementations on an FPGA.
The GNN-based approach allows for a flexible input scheme and architecture that can effectively consider many of these intricacies.
The theoretical advantage of using a graph structure as our input is that underlying traits can be derived from the structure of the input models, allowing commonalities to be found while not requiring overly specific engineering of the training data. 

Just as the benchmark and dataset are currently in their first iteration and set to evolve, we expect the surrogate model to see continued development beyond the scope of this work.
For an initial comparison, we set the baseline MLP from rule4ml against the more structured GNN and transformer implementations.


\subsection{Baseline MLP Implementation}
\label{subsec:benchmarkbaseline}

The baseline implementation uses a trained MLP model to predict each FPGA resource and latency variable. The general MLP architecture is similar to the one introduced in the open-source rule4ml tool, with minor adaptations to support our dataset. The architecture processes both numerical and categorical data extracted from an input model. First, ordinal encoding~\cite{potdar2017} is applied to categorical inputs, such as the target board and hls4ml strategy.
These encoded features are then processed through trainable embedding layers, which learn low-dimensional representations of the categorical data. Meanwhile, numerical inputs, composed mainly of statistical averages of the numerical features, are fed into a dense block, consisting of several fully connected layers with ReLU activations in between.
The outputs from this block are concatenated with the embeddings, creating a unified feature vector.
A final dense block processes the concatenated feature vector to produce the estimate.
Following the methodology in~\cite{rahali2024efficient}, an MLP model is trained per target variable for 200 epochs using the Adam optimizer~\cite{kingma2014Adam}, minimizing a mean squared logarithmic loss.

\subsection{Graph Neural Network}

Estimating the resource usage and latency of an arbitrary ML model presents unique challenges.
In particular, since a model can have any number of layers, each of which has arbitrary numbers of node connections, it is a challenge to effectively model these structures.
Furthermore, even relatively simple neural networks can have structures such as skip connections, which may have nontrivial effects on the resulting resource usage and latency of the inference engine.
Previous attempts to resolve this used the total number of layers of the network as a feature, along with other relevant values corresponding to the layers~\cite{rahimifar2024rule4ml}.
However, this may result in a severely limited scope of input models, since very different architectures may end up sharing nearly identical input features under these constraints, while their HLS syntheses produce very different circuits.
One way to handle these limitations would be to use a modeling structure with fewer limitations on how heterogeneous the input data can be.
For this, we take advantage of GNNs. Since the input to such a network can be an arbitrary graph, we can convert our input models into a graph representation, allowing for the heterogeneous data to be directly handled by our surrogate model.

\subsubsection{Features and Preprocessing}

Each layer of the input model is treated as a graph node with an 18-dimensional feature vector consisting of three input and output dimensions, precision, reuse factor, strategy, layer or activation type, filters, kernel size, stride, padding, batch normalization, and I/O type. Numerical features like layer dimensions and synthesis parameters are normalized via z-score standardization using training set statistics.
Categorical features, including layer type, activation, and padding, are one-hot encoded.

The graph is constructed by connecting nodes based on the sequential dataflow between layers.
Self-loops are added to each node to allow the attention mechanism to consider the layer's own features during message passing.
Global attributes that affect the entire model, such as synthesis strategy and I/O interface type, are also one-hot encoded and appended to each node's feature vector to provide consistent context.


\subsubsection{Structure}
\begin{figure}
 \centering
 \includegraphics[width=0.95\linewidth]{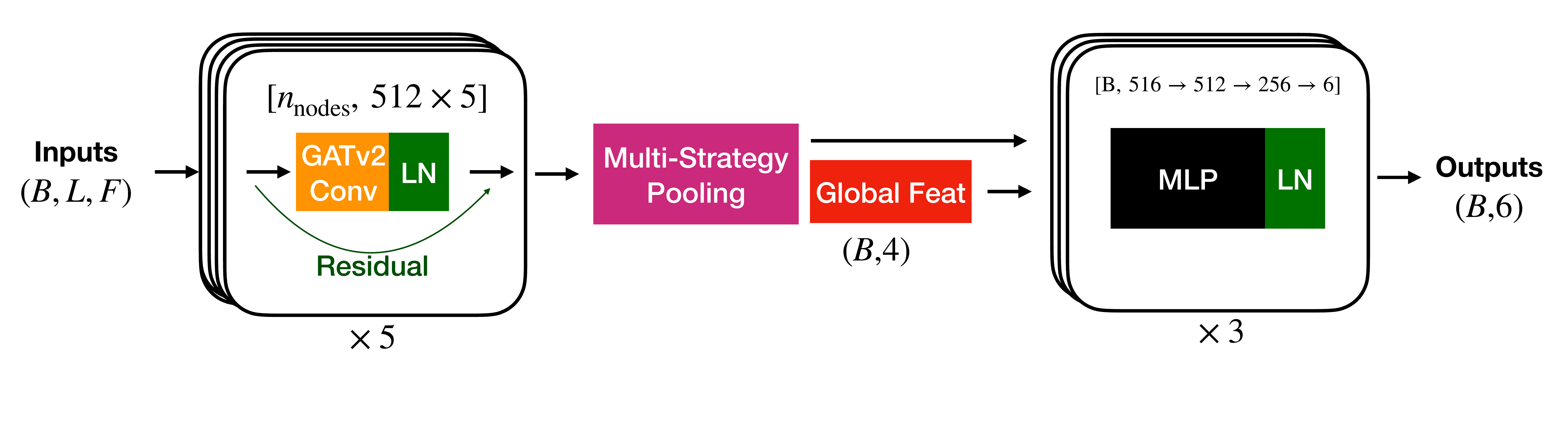}
 \caption{The overall structure of the GNN, comprising five GATv2Conv layers. The vector $(B, L, F)$ consists of $B$ batch size, $L$ layers per model, and $F$ features per layer.}
\label{gnnarch}
\end{figure}

The chosen GNN structure shown in \autoref{gnnarch} allows for arbitrary directed graph input, with each node having a fixed 18-dimensional feature vector after preprocessing.
It consists of five stacked graph attention network version 2 (GATv2) ~\cite{gatv1,gatv2,pytorch_geo} layers, each using five attention heads to capture various facets of inter-layer relationships.
The GATv2 attention mechanism assigns importance weights to edges dynamically based on the learned compatibility between node features.
The attention mechanism weighs edges according to their relative importance using the GATv2 formulation, where attention weights $\alpha_{i,j}$ are calculated as

\begin{equation}
    \mathbf{x}^{\prime}_i = \sigma\left(\sum_{j \in \mathcal{N}(i) \cup \{ i \}}
\alpha_{i,j}\mathbf{W}\mathbf{x}_{j}\right)
\end{equation}
where $\mathbf{W}$ is a trainable weight matrix for the linear transformation applied to the node features.
The attention coefficients, $\alpha_{i,j}$ which determine the importance of node \textit{j's} features to node \textit{i}, are computed dynamically for each edge using the GATv2 mechanism: 

\begin{equation}
\alpha_{i,j} = \frac{
\exp\left(\mathbf{a}^{\top}\mathrm{ELU}\left(
\mathbf{W}_{s} \mathbf{x}_i
+ \mathbf{W}_{t} \mathbf{x}_j
\right)\right)}
{\sum_{k \in \mathcal{N}(i) \cup \{ i \}}
\exp\left(\mathbf{a}^{\top}\mathrm{ELU}\left(
\mathbf{W}_{s} \mathbf{x}_i
+ \mathbf{W}_{t} \mathbf{x}_k
\right)\right)}
\end{equation}
where $\mathbf{a}$ is a learnable weight vector, and $\mathbf{W}_{s}$ and $\mathbf{W}_{t}$ are trainable weight matrices for the source and target nodes, respectively.

This allows the GNN to automatically determine which layer connections are most informative for prediction, even accounting for special structures like skip connections that may have an large impact on resource usage.
The GATv2 outputs undergo layer normalization, ELU activation, and dropout regularization, with residual connections to aid gradient flow in the deep architecture.

Node embeddings from the final GATv2 layer are first reduced to a standard size by a linear projection.
Then, a learnable weighted combination of additive, mean, and max pooling aggregates them into a single graph-level embedding, which is concatenated with the one-hot encoded global features.
This final representation passes through an MLP to yield the hardware usage estimates.


\subsubsection{Implementation}

We implemented the GNN using the PyTorch~\cite{pytorch} and PyTorch Geometric~\cite{pytorch_geo} libraries.
The JSON dataset was converted to NumPy arrays for efficient I/O, with logarithmic scaling applied to the target hardware metrics before z-score normalization to stabilize training.


The GNN architecture is built using standard modules from these libraries, including GATv2Conv or the attention-based graph convolutions and LayerNorm~\cite{ba2016layer} for stabilizing the activations between layers.
The model is trained by minimizing the mean-squared error (MSE) loss between the normalized predictions and targets using the AdamW optimizer ~\cite{loshchilov2019decoupled}.
We train the GNN network using our training set, which comprises 70\% of the total samples from the benchmark dataset.
A dynamic learning rate is employed, and the network trained for 200 epochs.
The training was performed with an NVIDIA A10 GPU.

\subsection{Transformer}

Similar to a GNN, a transformer architecture is a viable method to effectively estimate the resources and latency of models. With attention ~\cite{vaswani2023attentionneed}, the model complexity, scale, and relationship between layers can be better understood, by treating each layer as its own token.

\subsubsection{Features and Preprocessing}

As done for the GNN, the transformer follows a similar preprocessing procedure. The same 18-dimensional feature vector is produced, however no nodes are connected and no global features are created.

\subsubsection{Structure}

\begin{figure}
 \centering
 \includegraphics[width=1\linewidth]{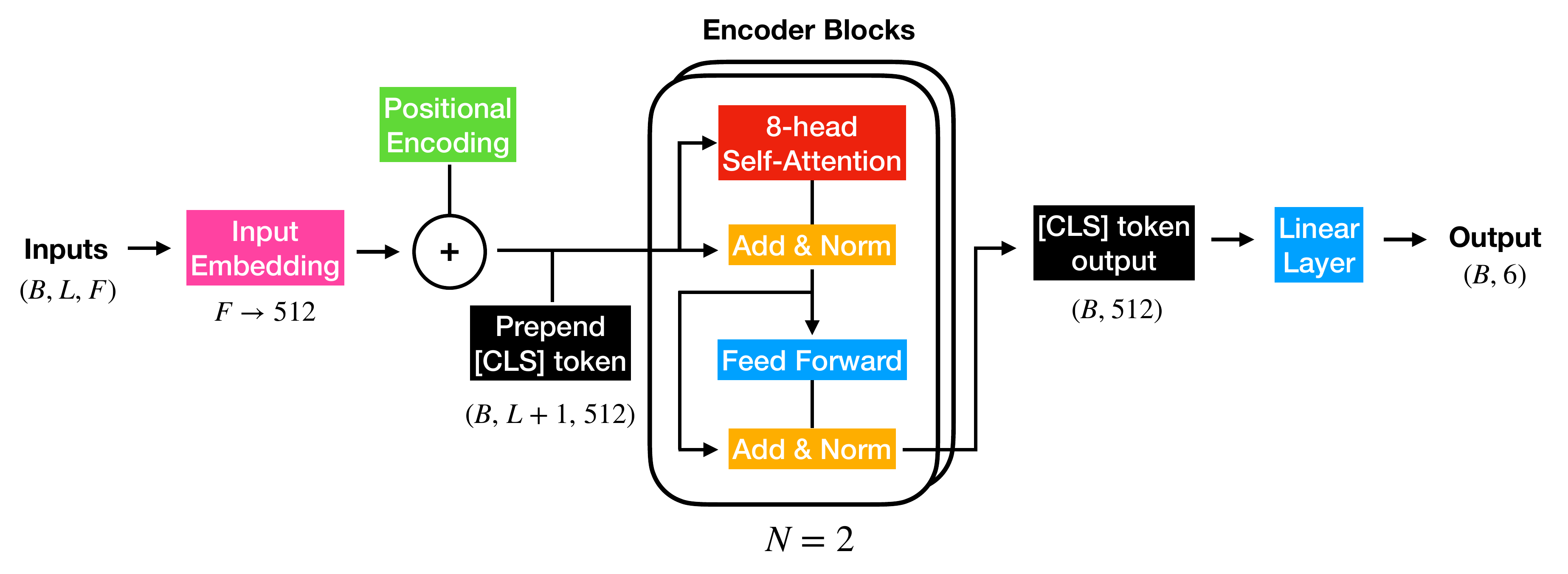}
 \caption{The overall structure of the transformer, comprising 2 encoder blocks. The vector $(B, L, F)$ consists of $B$ batch size, $L$ layers per model, and $F$ features per layer. $N=2$ denotes 2 sequential encoder blocks.}
\label{fig:transformerarchitecture}
\end{figure}

The transformer architecture is depicted in \autoref{fig:transformerarchitecture}.
The inputs are dimension $(B, L, F)$ where $B$ is the batch size, $L$ is the number of layers per model, and $F$ is the number of features per layer, which is 18.
The input embedding projects $F$ into a 512-dimensional embedding.
The positional encoding adds information about each layer's position relative to the others.
The [CLS] token is prepended to $L$, increasing the dimensionality by 1, aggregating the information of the entire sequence.
The two encoder blocks comprise an 8-head self-attention layer followed by a feed-forward network with normalization. The output of these blocks is the [CLS] token output, summarizing the whole model.
The linear layer maps 512 to 6 outputs, corresponding to the hardware resource predictions.

\subsubsection{Implementation}

The transformer was implemented using PyTorch~\cite{pytorch} for all model components.
Each input model is encoded as a sequence of features per layer, with padding applied up to a maximum of 51 layers.
All features are normalized based on the training set's mean and standard deviation.
A padding mask is generated for each sample to indicate which layers are real or padded.

Each feature vector per layer of length 18 is projected into a 512-dimensional token embedding by a learned linear layer.
Learnable positional encodings are added to each token to encode the order of layers.
A [CLS] token is prepended to each sequence.
Padding masks are passed into the transformer to prevent attention to padded tokens.

The architecture, as defined above, uses nn.TransformerEncoder and nn.TransformerEncoderLayer.
Dropout is applied within the encoder layers for regularization.
The output predicts the 6 hardware metrics.

The model is trained for 250 epochs with a batch size of 1024.
Training is performed on an NVIDIA A100 GPU.
For inference, the outputs are rescaled to the original metrics scales for accurate predictions.
\section{Results}
\subsection{Relative Percent Error}
\begin{figure}
\centering
\includegraphics[width=\linewidth, page=1]{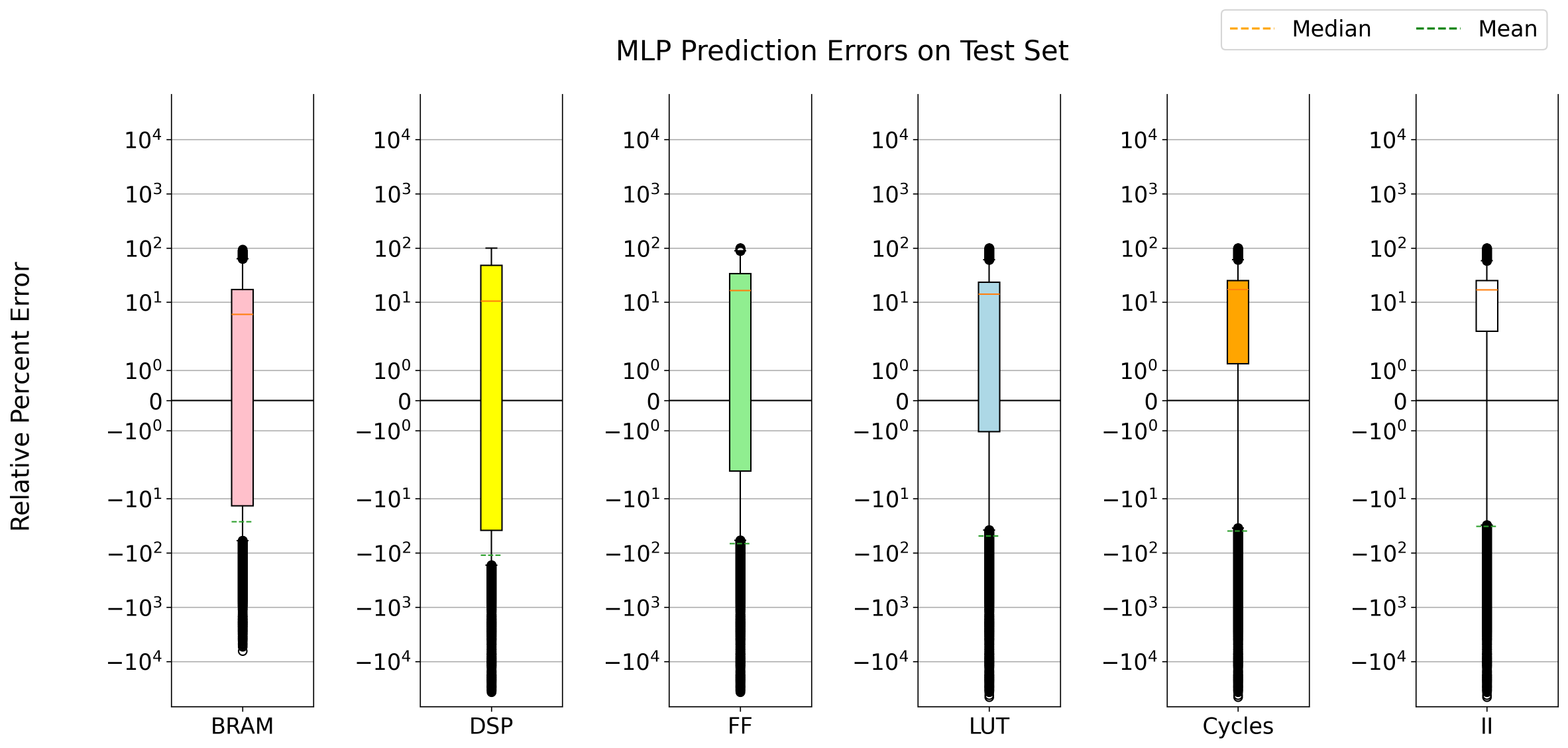}
\caption{Relative percentage errors of the baseline MLP on the test set. The y-axis is set to a symmetric log scale.}
\label{fig:mlp-box-test}
\end{figure}

\begin{figure}
\centering
\includegraphics[width=\linewidth, page=2]{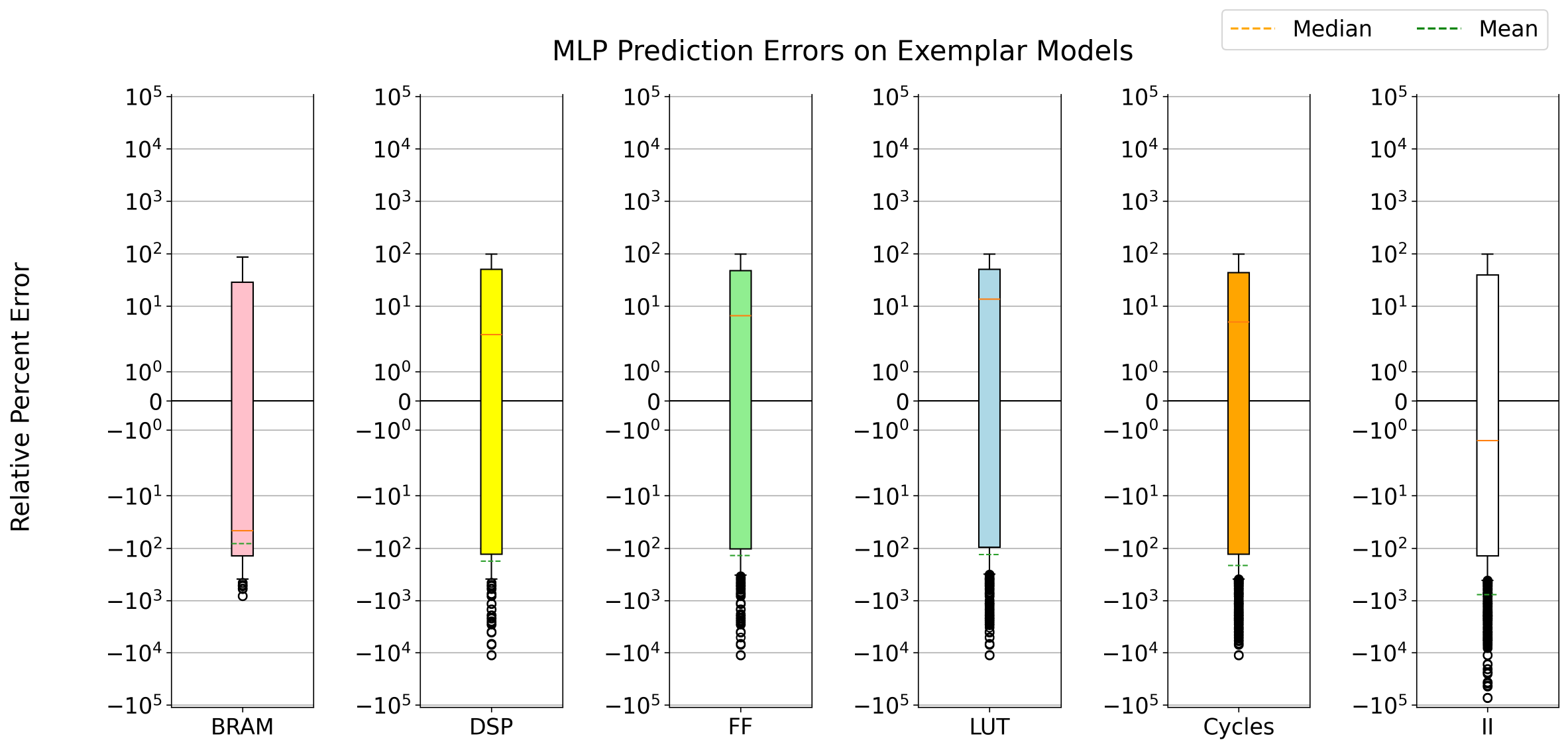}
\caption{Relative percentage errors of the MLP on the exemplar models. The y-axis is set to a symmetric log scale.}
\label{fig:mlp-box-benchmark}
\end{figure}

\begin{figure}
 \centering
 \includegraphics[width=\linewidth]{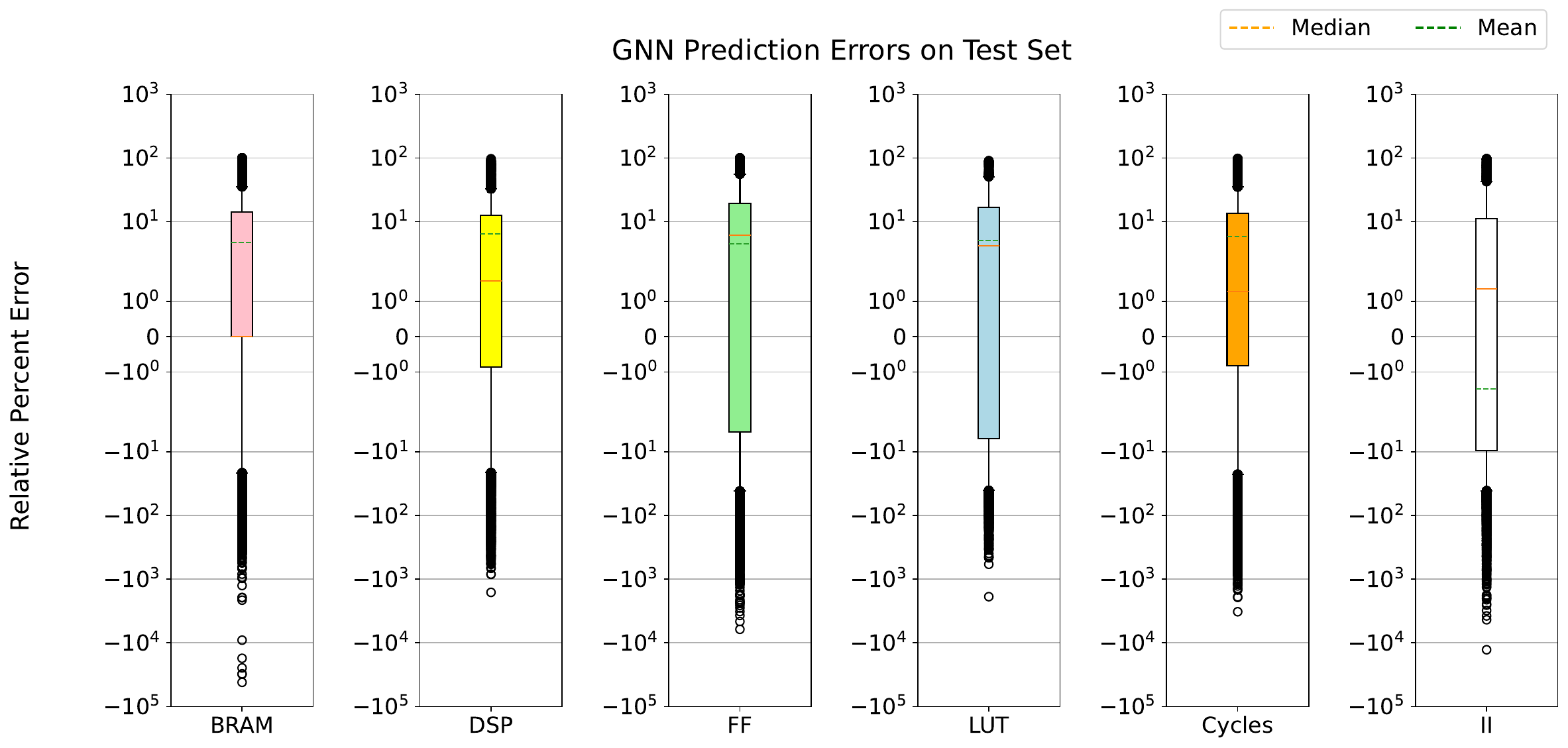}
 \caption{Relative percent error for the GNN on the test subset. The y-axis is set to a symmetric log scale.}
 \label{fig:GNN-test}
\end{figure}

\begin{figure}
 \centering
 \includegraphics[width=\linewidth]{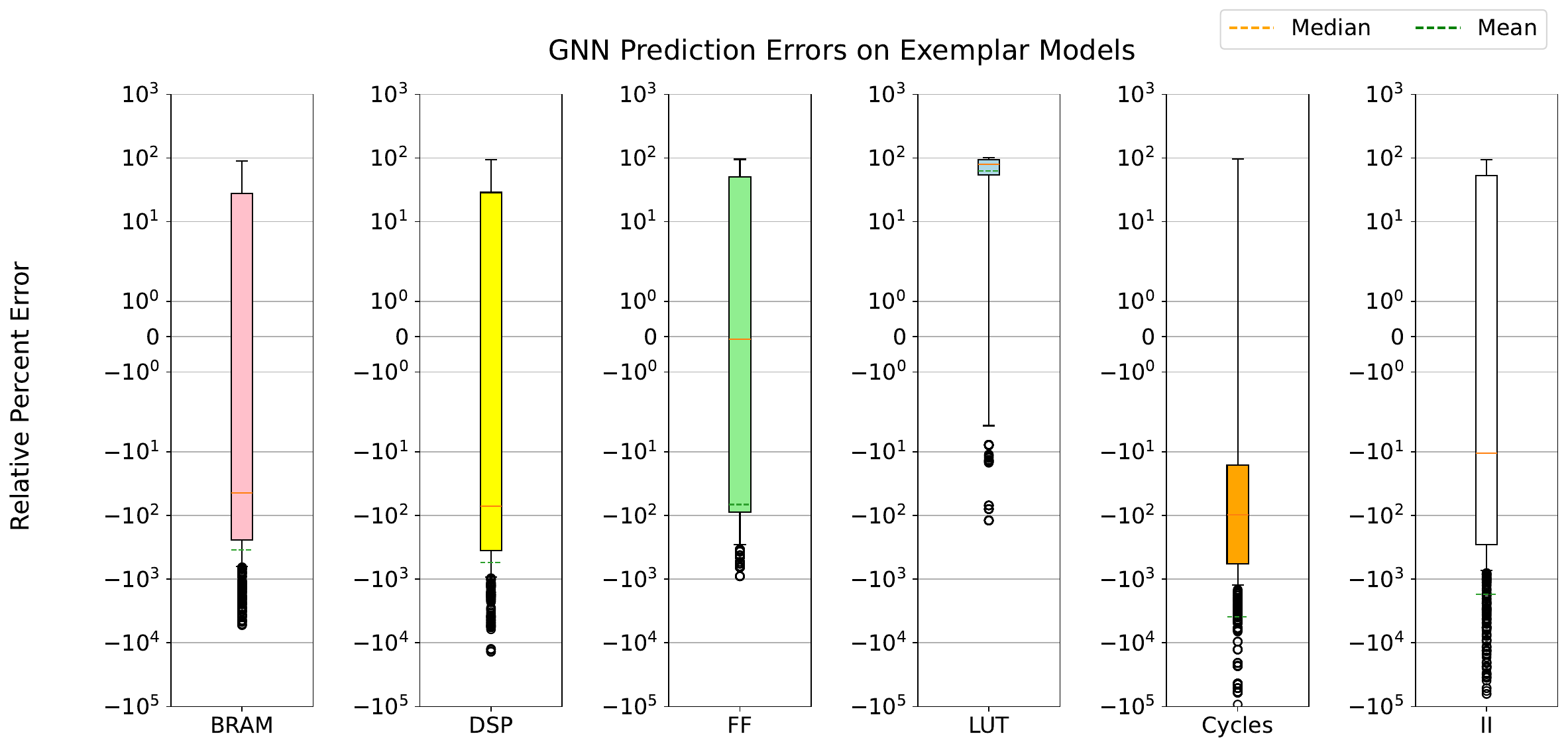}
 \caption{Relative percent error for the GNN on the exemplar dataset. The y-axis is set to a symmetric log scale. Activation layers are removed from the exemplar set to keep a similar input structure as the GNN was trained on.}
 \label{fig:GNN-exemplar}
\end{figure}

\begin{figure}
 \centering
 \includegraphics[width=\linewidth]{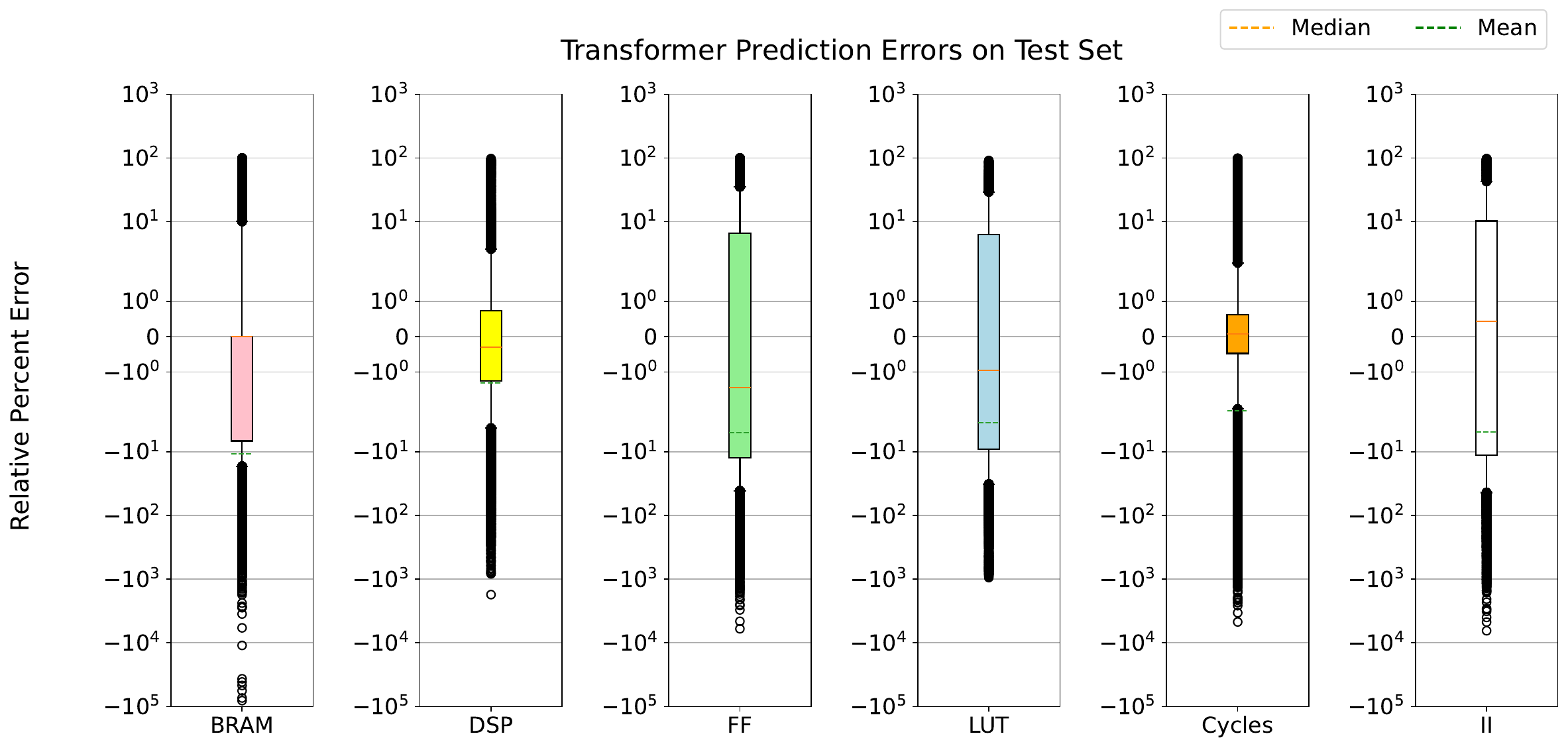}
 \caption{Relative percent error for the transformer on the test subset. The y-axis is set to a symmetric log scale.}
 \label{fig:transformer-test}
\end{figure}

\begin{figure}
 \centering
 \includegraphics[width=\linewidth]{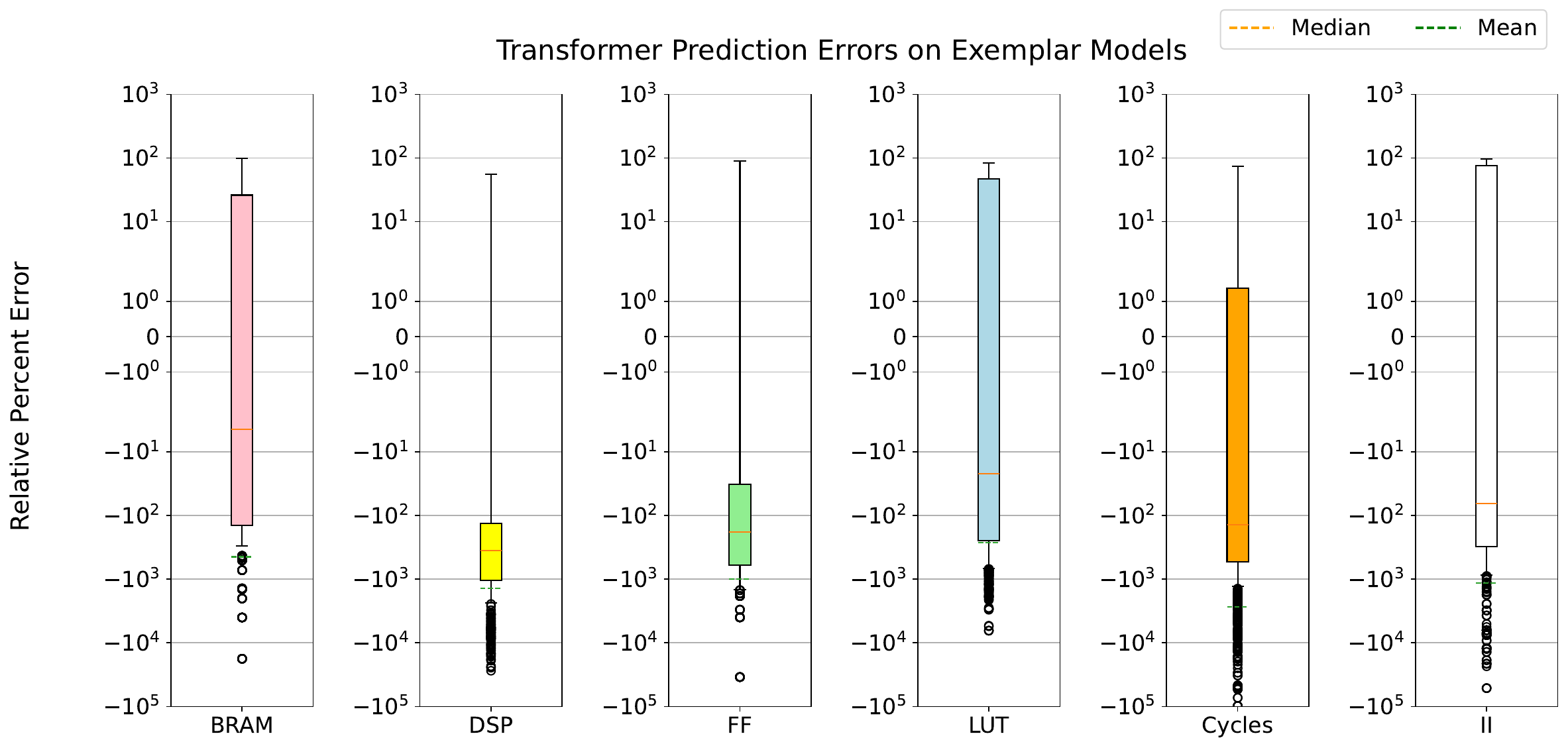}
 \caption{Relative percent error for the transformer on the exemplar subset. The y-axis is set to a symmetric log scale.}
 \label{fig:transformer-exemplar}
\end{figure}

As discussed in \autoref{subsec:benchmark_metrics}, we visualize results with the RPE, relative percent error, using box plots.
The results are shown for each resource target (BRAM, DSP, FF, and LUT) and latency target (clock cycles) and initiation interval (II) in a box plot.
We present results for the synthetic model test samples and exemplar realistic models for the baseline MLP, GNN, and transformer prediction models. 
\autoref{fig:mlp-box-test} and~\autoref{fig:mlp-box-benchmark} show the synthetic model test set and the exemplar models, respectively, for the baseline MLP. 
\autoref{fig:GNN-test} and~\autoref{fig:GNN-exemplar} show the synthetic model test set and the exemplar models, respectively, for the GNN model. \autoref{fig:transformer-test} and~\autoref{fig:transformer-exemplar} show the synthetic model test set and the exemplar models, respectively, for the transformer model.
In each plot, the colored box covers the interquartile range (IQR), spanning both the first quartile (25\%) and the third (75\%). The dashed horizontal lines within the box show the median (orange) and mean (green) of the distribution. The whiskers extend from the box to the smallest and largest values within 1.5 times the IQR, capturing most of the data spread and considering points outside this range as outliers.

\begin{table*}
\footnotesize
\renewcommand{\arraystretch}{1.3}
\setlength\tabcolsep{1.6pt}
\centering
\caption{Evaluation metrics and results of the surrogate models on the test set and its subsets.}
\label{tab:metrics-test}
\resizebox{\textwidth}{!}{
\begin{tabular}{P{0.4in}|c|cccccc|cccccc|cccccc|}
\multirow{2}{*}{Models} & \multirow{2}{*}{Arch.} & \multicolumn{6}{c|}{$R^2$ Score} & \multicolumn{6}{c|}{SMAPE [\%]} & \multicolumn{6}{c|}{RMSE} \\ 
\cline{3-20} 
    & & BRAM & DSP & FF & LUT & Cycles & II & BRAM & DSP & FF & LUT & Cycles & II & BRAM & DSP & FF & LUT & Cycles & II \\
\hline \hline
\multirow{3}{*}{\shortstack{Test set\\(All)}}
& MLP
& 0.32 & 0.03 & 0.20 & 0.49 & 0.54 & 0.56
& 33.9 & 105.7 & 24.9 & 15.5 & 31.8 & 25.8
& \textbf{12.5} & 590.4 & \textbf{31897.9} & \textbf{40463.1} & 450879.5 & 221524.7 \\

& GNN
& \textbf{0.51} & \textbf{0.89} & 0.74 & \textbf{0.73} & 0.89 & 0.91
& 19.5 & 15.1 & 11.6 & 11.4 & 15.7 & \textbf{13.4}
& 48.0 & \textbf{580.0} & 55087.6 & 104945.7 & 227987.8 & 201369.9 \\

& Transformer
& 0.39 & 0.29 & \textbf{0.72} & 0.67 & \textbf{0.95} & \textbf{0.95}
& \textbf{14.1} & \textbf{10.8} & \textbf{2.9} & \textbf{2.9} & \textbf{10.1} & 14.1
& 53.7 & 1472.3 & 57138.2 & 115943.6 & \textbf{147137.7} & \textbf{150688.4} \\

\hline
\multirow{3}{*}{\shortstack{Test set\\(Dense)}}
& MLP
& \textbf{0.47} & 0.03 & 0.13 & 0.49 & 0.74 & 0.77
& 33.3 & 106.6 & 24.5 & 14.8 & 30.9 & 24.8
& \textbf{9.3} & \textbf{603.0} & \textbf{31441.0} & \textbf{41221.3} & 1548.5 & 639.8 \\

& GNN
& -0.51 & -0.74 & \textbf{0.73} & \textbf{0.73} & 0.82 & \textbf{0.91}
& 24.6 & 23.9 & 11.8 & 11.6 & 15.8 & \textbf{13.4}
& 86.1 & 18950.8 & 56341.8 & 107544.5 & 1304.9 & 415.1 \\

& Transformer
& 0.39 & \textbf{0.29} & 0.71 & 0.67 & \textbf{0.95} & \textbf{0.91}
& \textbf{13.6} & \textbf{10.2} & \textbf{2.7} & \textbf{2.7} & \textbf{9.9} & 14.1
& 54.9 & 1509.2 & 58463.5 & 118826.6 & \textbf{659.7} & \textbf{395.6} \\

\hline
\multirow{3}{*}{\shortstack{Test set\\(Conv1D)}}
& MLP
& 0.39 & -0.07 & 0.31 & 0.24 & 0.38 & 0.40
& 44.7 & 80.5 & 28.7 & 27.2 & 48.4 & 43.8
& \textbf{8.9} & 2.5 & 14249.3 & 8994.6 & 98637.3 & 48849.3 \\

& GNN
& 0.69 & 0.02 & 0.95 & \textbf{0.96} & \textbf{0.97} & \textbf{0.97}
& 33.7 & 36.7 & 7.7 & 6.2 & 11.0 & \textbf{11.1}
& 16.9 & 1.7 & 5137.6 & 3731.4 & \textbf{21648.4} & \textbf{24199.6} \\

& Transformer
& \textbf{0.77}   & \textbf{0.41}  & \textbf{0.97}   & \textbf{0.96}    & 0.96   & 0.96
& \textbf{29.9}  & \textbf{22.7} & \textbf{7.2}   & \textbf{5.8}    & \textbf{10.4}  & 11.2
& 14.5  & \textbf{1.3}  & \textbf{4275.9} & \textbf{3421.7} & 24561.6 & 27547.3 \\

\hline
\multirow{3}{*}{\shortstack{Test set\\(Conv2D)}}
& MLP
& -0.50 & 0.15 & 0.51 & 0.41 & 0.33 & 0.35
& 51.5 & 87.0 & 41.8 & 34.4 & 56.1 & 54.7
& 59.6 & 6.0 & 57190.7 & 18547.8 & 3181122.1 & 1562928.9 \\

& GNN
& 0.44 & 0.51 & 0.92 & 0.95 & 0.84 & 0.88
& 31.2 & 34.8 & 8.8 & \textbf{6.7} & 19.5 & 18.5
& 27.4 & 5.5 & 24145.3 & 15283.9 & 1549792.9 & 1365427.3 \\

& Transformer
& \textbf{0.79}   &\textbf{0.55}  & \textbf{0.93}   & \textbf{0.96}    & \textbf{0.93}    & \textbf{0.93}
& \textbf{18.8}  & \textbf{25.3} & \textbf{8.0}   & 6.8    & \textbf{16.3}   & \textbf{16.9}
& \textbf{16.7}  & \textbf{5.3}  & \textbf{22659.7} & \textbf{12971.7}  & \textbf{999953.4} & \textbf{1024014.3} \\

\end{tabular}}
\end{table*}

\begin{table*}
\renewcommand{\arraystretch}{1.3}
\setlength\tabcolsep{1.6pt}
\centering
\caption{Evaluation metrics and results of the surrogate models on the exemplar architectures.}
\label{tab:metrics-examplar}
\resizebox{\textwidth}{!}{
\begin{tabular}{P{0.5in}|c|cccccc|cccccc|cccccc|}
\multirow{2}{*}{Models} & \multirow{2}{*}{Arch.} & \multicolumn{6}{c|}{$R^2$ Score} & \multicolumn{6}{c|}{SMAPE [\%]} & \multicolumn{6}{c|}{RMSE} \\ 
\cline{3-20} 
    & & BRAM & DSP & FF & LUT & Cycles & II & BRAM & DSP & FF & LUT & Cycles & II & BRAM & DSP & FF & LUT & Cycles & II \\
\hline \hline

\multirow{3}{*}{Jet}
& MLP
& \textbf{-1.22} & 0.27 & 0.33 & -0.12 & \textbf{0.53} & \textbf{0.49}
& \textbf{65.0} & 86.0 & \textbf{58.8} & \textbf{74.1} & 89.1 & \textbf{66.4}
& \textbf{2.2} & 569.8 & 8654.0 & 18214.8 & \textbf{711.4} & \textbf{393.6} \\

& GNN
& -1.29 & -0.14 & 0.12 & \textbf{0.19} & 0.43 & 0.41
& 143.8 & 170.3 & 86.3 & 80.8 & \textbf{86.6} & 103.6
& 2.3 & 712.2 & 9917.7 & \textbf{15516.7} & 780.3 & 424.2 \\

& Transformer
& -23.19 & \textbf{0.32} & \textbf{0.46} & -0.03 & 0.17 & 0.04
& 110.3 & \textbf{77.7} & 80.6 & 90.1 & 103.4 & 120.4
& 7.4 & \textbf{550.8} & \textbf{7764.8} & 17520.1 & 944.2 & 542.0 \\

\hline
\multirow{3}{*}{Quarks}
& MLP
& N/A\textsuperscript{*} & \textbf{0.19} & \textbf{0.51} & \textbf{-0.41} & -36.68 & -13.18
& \textbf{129.9} & \textbf{118.1} & \textbf{83.5} & \textbf{91.9} & 158.4 & \textbf{136.3}
& 2.3 & 108.6 & \textbf{1327.1} & \textbf{3298.1} & 441.1 & 266.6 \\

& GNN
& N/A\textsuperscript{*} & -0.31 & -0.64 & -11.33 & -95.13 & -33.73
& 200.0 & 119.9 & 105.8 & 113.6 & 171.1 & 170.2
& \textbf{1.2} & 137.9 & 2432.5 & 9756.8 & 704.5 & 417.3 \\

& Transformer
& N/A\textsuperscript{*} & -0.28 & 0.24 & -12.46 & \textbf{-4.21} & \textbf{-2.61}
& 200.0 & 138.4 & 92.0 & 117.4 & \textbf{144.2} & 152.5
& 119.9 & \textbf{136.2} & 1655.9 & 10193.2 & \textbf{164.0} & \textbf{134.5} \\

\hline
\multirow{3}{*}{Anomaly}
& MLP
& -0.80 & 0.26 & \textbf{0.59} & \textbf{0.45} & 0.42 & \textbf{0.49}
& \textbf{104.3} & \textbf{59.2} & \textbf{36.9} & \textbf{49.2} & 51.5 & 107.9
& 4.4 & 500.7 & \textbf{12774.2} & \textbf{14251.0} & 761.2 & \textbf{376.7} \\

& GNN
& \textbf{-0.64} & 0.17 & -0.54 & -6.91 & \textbf{0.43} & 0.46
& 117.5 & 185.8 & 74.8 & 91.5 & 76.2 & \textbf{86.7}
& \textbf{4.2} & 531.8 & 24620.3 & 53979.5 & \textbf{755.7} & 384.6 \\

& Transformer
& -185.38 & \textbf{0.71} & -10.84 & -135.03 & 0.32 & 0.32
& 168.9 & 72.3 & 121.3 & 160.7 & \textbf{45.8} & 117.3
& 44.8 & \textbf{312.1} & 68319.0 & 223829.6 & 820.9 & 434.9 \\

\hline
\multirow{3}{*}{BiPC}
& MLP
& \textbf{-0.71} & -0.04 & \textbf{0.16} & \textbf{0.03} & \textbf{0.44} & 0.43
& \textbf{107.7} & 127.0 & \textbf{77.2} & \textbf{82.3} & 75.9 & \textbf{70.6}
& \textbf{3.4} & 1719.2 & \textbf{24472.0} & \textbf{48867.4} & \textbf{1821.4} & 573.6 \\

& GNN
& -0.93 & -0.10 & -0.35 & -0.24 & 0.26 & \textbf{0.45}
& 145.2 & \textbf{118.5} & 111.5 & 89.6 & 87.7 & 119.1
& 3.6 & 1776.3 & 30980.1 & 55222.2 & 2090.1 & \textbf{559.7} \\

& Transformer
& -17.38  & \textbf{0.12}   & -16.70  & -12.16   & 0.43    & \textbf{0.45}
& 135.1  & 136.2 & 124.6  & 120.4   & \textbf{69.3}   & 101.3
& 11.1   & \textbf{1588.1} & 112115.9 & 179918.2 & 1832.6 & 563.7 \\

\hline
\multirow{3}{*}{\shortstack{Cookie-\\box}}
& MLP
& \textbf{-0.54} & 0.20 & \textbf{0.34} & \textbf{0.13} & \textbf{0.45} & \textbf{0.53}
& \textbf{64.4} & 91.3 & \textbf{66.8} & \textbf{67.9} & \textbf{32.2} & \textbf{72.3}
& \textbf{3.5} & 657.9 & 11675.1 & \textbf{21420.7} & \textbf{653.5} & \textbf{341.0} \\

& GNN
& -19.23 & \textbf{0.98} & -1.92 & -90.18 & 0.21 & 0.26
& 157.2 & 137.6 & 114.4 & 118.4 & 45.7 & 82.6
& 12.7 & \textbf{104.5} & 24559.3 & 219369.3 & 784.5 & 429.1 \\

& Transformer
& -37.91  & 0.64   & \textbf{0.34}    & -19.36    & 0.26   & -0.02
& 141.9  & \textbf{88.0}  & 93.3   & 138.4    & 39.0  & 130.7
& 17.6   & 438.8 & \textbf{11657.6} & 103672.4 & 762.4 & 503.6  \\

\hline
\multirow{3}{*}{AutoMLP}
& MLP
& -1.09 & \textbf{0.41} & \textbf{0.69} & \textbf{-0.22} & -0.33 & -1.71
& \textbf{56.4} & \textbf{72.1} & \textbf{61.4} & \textbf{68.9} & 86.8 & \textbf{86.0}
& 0.9 & \textbf{104.4} & \textbf{1524.8} & \textbf{3586.5} & 226.5 & 163.0 \\

& GNN
& \textbf{-1.04} & -0.15 & 0.19 & -0.56 & -0.08 & -0.34
& 161.8 & 180.8 & 79.9 & 78.9 & \textbf{79.0} & 96.2
& \textbf{0.9} & 145.1 & 2459.0 & 4056.9 & \textbf{205.6} & 114.7 \\

& Transformer
& -59.19  & 0.01   & 0.15    & -13.91    & \textbf{0.20}    & \textbf{0.22}
& 118.1  & 95.4  & 74.2   & 120.8    & 101.8  & 113.9
& 4.9    & 134.6 & 2528.0 & 12540.5  & 177.1  & \textbf{87.5}   \\

\hline
\multirow{3}{*}{\shortstack{Particle\\Tracking}}
& MLP
& -1.41 & 0.28 & 0.33 & -0.08 & \textbf{0.52} & \textbf{0.50}
& \textbf{65.6} & \textbf{75.2} & \textbf{58.1} & \textbf{71.2} & 87.5 & \textbf{61.6}
& 2.1 & 536.1 & 8093.3 & 16695.1 & \textbf{692.9} & \textbf{382.0} \\

& GNN
& \textbf{-1.03} & -0.13 & 0.19 & \textbf{0.14} & 0.45 & 0.40
& 144.4 & 158.5 & 84.8 & 81.4 & \textbf{83.5} & 100.9
& \textbf{2.0} & 670.2 & 9100.7 & \textbf{15013.8} & 745.5 & 419.1 \\

& Transformer
& -24.92  & \textbf{0.34}   & \textbf{0.47}    & -0.01     & 0.15    & 0.04
& 118.5  & 83.7  & 80.6   & 89.1     & 108.0  & 124.5
& 7.0    & \textbf{512.5} & \textbf{7384.7} & 16204.8  & 927.1  & 529.5  \\

\end{tabular}}
\raggedright \textsuperscript{*} {\footnotesize $R^2$ score calculation is skipped since all true values are 0.} \\
\end{table*}

Based on the test set RPE boxplots in \autoref{fig:GNN-test} and \autoref{fig:transformer-test}, the GNN and transformer models demonstrate a significant improvement over the baseline MLP. As shown in \autoref{fig:mlp-box-test}, the MLP predictions generally have wider interquartile ranges (IQRs) and medians. The transformer model, in particular, shows very narrow IQRs for DSP and Cycles near zero. The GNN tends to have somewhat narrow IQRs for BRAM, DSP, and Cycles, also having a tendency to over-predict for these features, which may be a more desirable behavior than under-predicting. 

When evaluated on the exemplar dataset, all three predictors show a drop in performance, highlighting the challenge of generalizing new and complex architectures not present in the training data. The RPE for the exemplar set, illustrated in \autoref{fig:mlp-box-benchmark}, \autoref{fig:GNN-exemplar}, and \autoref{fig:transformer-exemplar}, shows considerably wider and considerably more outliers compared to the test set results. Looking at the median and mean, the MLP tends to over-predict, whereas the GNN and transformer under-predict, likely due to their log scaling pre-processing step.

\subsection{Full evaluation metrics}
For a quantitative analysis, we evaluated the $R^2$, SMAPE, and RMSE metrics defined in \autoref{subsec:benchmark_metrics} for each surrogate model.
These metrics were computed for the test set and each architecture within the exemplar set.
The results are summarized in \autoref{tab:metrics-test} for the test set and \autoref{tab:metrics-examplar} for the exemplar set. 

On the test set, \autoref{tab:metrics-test}, the three models show distinct performance patterns. The transformer has the highest $R^2$ scores for Cycles (0.95) and II (0.95), while the GNN performs best for DSP (0.89) and competitive scores for other resources. The MLP shows lower $R^2$ values across most metrics, particularly for DSP (0.03). In terms of SMAPE, the transformer achieves lower errors for FF (2.9\%) and LUT (2.9\%), while the GNN shows the best performance for II (13.4\%). RMSE results vary considerably, with the MLP showing the lowest values for BRAM, FF, and LUT, though this may reflect its tendency towards smaller absolute predictions rather than better accuracy.

Performance varies significantly across the layer types of the test set. For dense layers, all models show improved performance compared to the overall set, likely due to their relative simplicity and high representation in the overall dataset. For Conv1D layers, both the GNN and transformer show strong prediction with $R^2$ values exceeding 0.95 for most metrics, while the MLP struggles particularly with DSP ($R^2 = -0.7$). Conv2D layers present the greatest challenge, though the transformer still achieves $R^2$ values above 0.90 for Cycles and II.

The exemplar architectures \autoref{tab:metrics-examplar} highlight the generalization challenges faced by all models. Negative $R^2$ values are common across all models, indicating predictions worse than the mean. The MLP shows negative $R^2$ for BRAM in most cases but performs relatively better for Cycles and II in some architectures (Jet: 0.53, 0.49). The GNN shows mixed performance, with particularly poor performance on Quarks and Anomaly architectures but reasonable performance for specific metrics in other cases. The transformer shows the most variability, achieving the best DSP predictions for several architectures but struggling with BRAM predictions.

SMAPE values for the exemplar are also substantially higher than the test set across all models. The transformer generally achieves lower SMPAPE values for resource utilization (particularly DSP), while showing competitive performance for timing metrics. 


As seen in \autoref{sec:scatter_plots}, all models are better at predicting resources and latency for the test set compared to the exemplar set. Relative to the MLP, the GNN and transformer are able to consistently predict DSP and Cycles on the test set.
Overall, the discrepancy between the exemplar and test data performance we see among all models can be attributed to the distribution shown in \autoref{fig:exemplar-label-distribution}, where the exemplar data is not reflected by the training and test subsets. This highlights the need to further improve the diversity of the dataset to include a wider variety of model architectures.

\section{Summary and Outlook}
We developed wa-hls4ml as a benchmark to provide a standardized method to evaluate the performance of neural-network-based FPGA resource estimation tools.
Alongside the associated dataset, we hope to provide a comprehensive performance evaluation scheme and a basis for the further development of similar tools, in line with previous benchmark efforts~\cite{borras2022open,duarte2022fastmlscience}. 

In presenting the GNN and transformer-based neural networks, we seek to demonstrate the performance of novel architectures for estimating latency and resource utilization of neural networks on FPGAs through hls4ml.
The results demonstrate that the surrogate models perform well on the test dataset, indicating that our approach toward estimating resources and latency is viable and that further research into these methods is warranted. 

While the estimators perform well on test data similar to their training set, their performance on the realistic exemplar set, which contains varying architectures and different configurations not present in the training data, is lacking.
The ability of the GNN and transformer to handle such varied architectures implicitly still offers great potential, especially with the development of a more robust dataset. 

This work demonstrates one concrete application where the surrogate model rapidly predicts resource/latency estimates, but the wa-hls4ml dataset is intentionally broader. Including the full project for each synthesized model (HLS code, IR and multi-stage reports) opens up other applications such as code/IR-driven learning (code2vec embeddings, code autocomplete), budget-aware architecture recommendation, and LLM-based assistants tailored to HLS.

In future work, we intend to expand the provided dataset to include larger neural networks, more intricate architectures that include features like skip connections, a larger variety of hardware settings like reuse factors, and a larger number of samples in the dataset.
As hls4ml evolves to support more architectures, we intend to continuously extend the dataset to reflect the added features.

We additionally intend to further improve the GNN and transformer models not only with an improved training dataset, but through further refinements of their architecture, and the exploration of techniques such as preferring overestimation when calculating the loss.
Similarly, we intend to continue to develop and update the benchmark periodically, incorporating new metrics, tracked features, and dataset improvements as appropriate. 

\section{Dataset and Code Availability}
\label{sec:data_code}

The training, test, and exemplar test datasets discussed in this work are available here: \url{https://huggingface.co/datasets/fastmachinelearning/wa-hls4ml}, licensed under the CC-BY-NC 4.0 license.
Additionally, a dataset containing the corresponding synthesized project files and logs for the samples in the training, test, and exemplar datasets is available here: \url{https://huggingface.co/datasets/fastmachinelearning/wa-hls4ml-projects}, licensed under the CC-BY-NC 4.0 license.
The code used to generate the datasets, plots, models, and other associated results are available at the following meta repository, licensed under the respective licenses as mentioned in the repositories: \url{https://github.com/fastmachinelearning/wa-hls4ml-paper}.
\begin{acks}
This manuscript has been authored by FermiForward Discovery Group, LLC under Contract No. 89243024CSC000002 with the U.S. Department of Energy, Office of Science, Office of High Energy Physics.

This work was supported in part by the U.S. Department of Energy, Office of Science, Office of Workforce Development for Teachers and Scientists (WDTS) under the Science Undergraduate Laboratory Internships Program (SULI).

Compute was provided in part by the Elastic Analysis Facility at Fermilab. 

Portions of this research were conducted with the advanced computing resources provided by Texas A\&M High Performance Research Computing.

This work was supported in part by National Science Foundation (NSF) awards CNS-1730158, ACI-1540112, ACI-1541349, OAC-1826967, OAC-2112167, CNS-2100237, CNS-2120019, the University of California Office of the President, and the University of California San Diego's California Institute for Telecommunications and Information Technology/Qualcomm Institute.
Thanks to CENIC for the 100 Gbps networks.

KS is supported by National Science Foundation (NSF) Grants 2112356 and 2411377

BH, DP, KT, GG, and NT are supported by Fermi Research Alliance, LLC under Contract No. DE-AC02-07CH11359 with the United States Department of Energy (DOE), Office of Science, Office of High Energy Physics. BH and NT are also supported under the DOE Early Career Research program under Award No. DE-0000247070. KT is also supported by DOE Grant KA2401045. BH, JD, and NT are supported by the U.S. Department of Energy (DOE), Office of Science, Office of Advanced Scientific Computing Research under the ``Real-time Data Reduction Codesign at the Extreme Edge for Science'' Project(DE-FOA-0002501).

JD is supported by the Research Corporation for Science Advancement (RCSA) under grant \#CS-CSA-2023-109, Alfred P. Sloan Foundation under grant \#FG-2023-20452, U.S. Department of Energy (DOE), Office of Science, Office of High Energy Physics Early Career Research program under Award No. DE-SC0021187, and the U.S. National Science Foundation (NSF) Harnessing the Data Revolution (HDR) Institute for Accelerating AI Algorithms for Data Driven Discovery (A3D3) under Cooperative Agreement PHY-2117997.

Thank you to Prof. Luca Carloni at Columbia University for for supporting the work of DS through the CSEE–E6868: Embedded Scalable Platforms – Spring ‘25 course.

JW is supported by a WATCHEP fellowship sponsored by the DOE, Office of High-Energy Physics under Award No. DE-SC-0023527.

HER, MMR, and ACT are supported by funding from the Canada Research Chairs Program. ACT holds the Canada Research Chair in Real-Time Intelligence Embedded for High-Speed Sensors.
\end{acks}

\bibliographystyle{ACM-Reference-Format}
\bibliography{References}

\clearpage

\appendix
\begin{appendices}
\section{Scatter Plots}
\label{sec:scatter_plots}
\begin{figure}[htb]
\centering
\includegraphics[width=\linewidth, page=1]{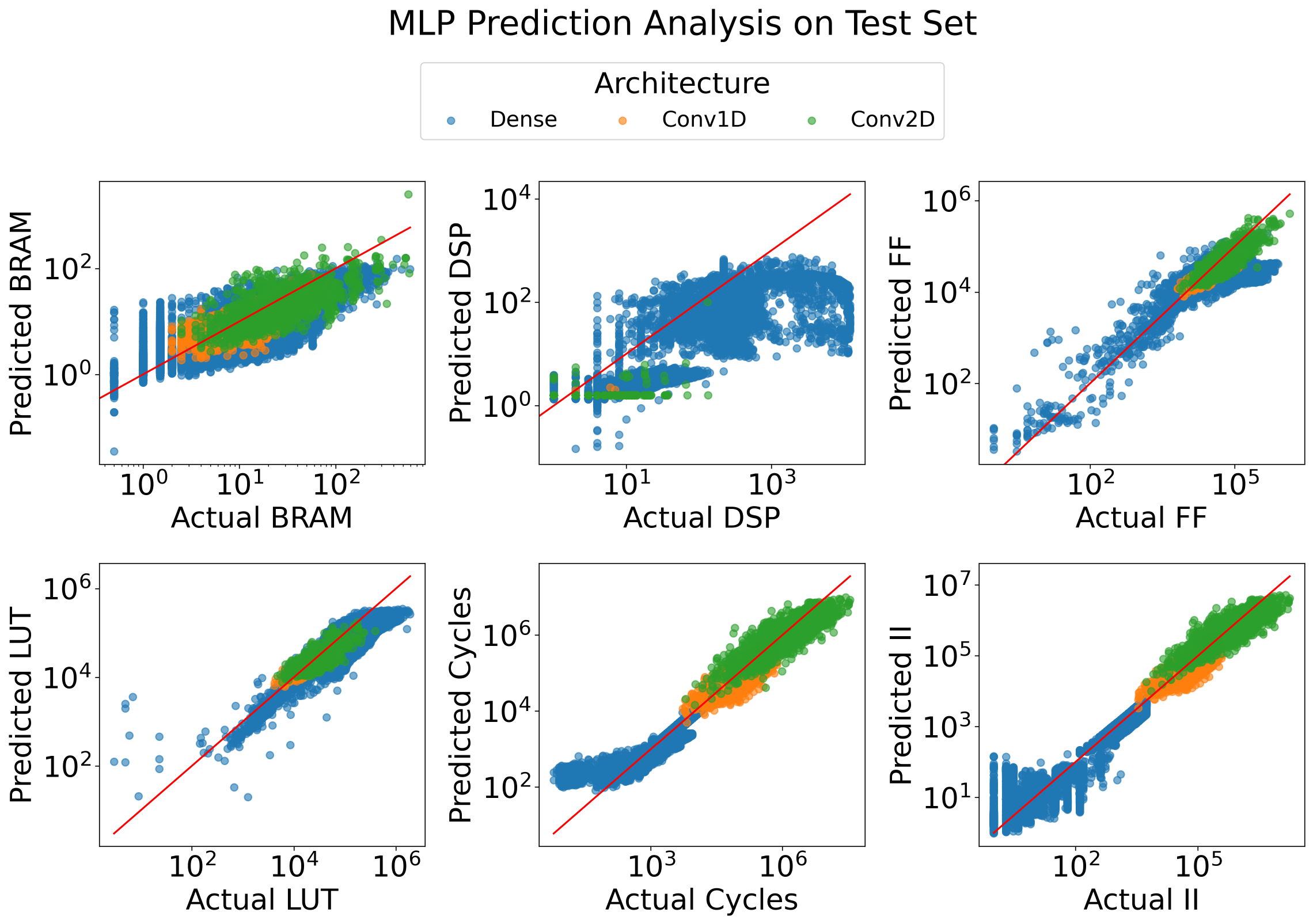}
\caption{Scatter plots of MLP predictions on the test set. The red line shows the deviation from true values. Both axes are set to a logarithmic scale.}
\label{fig:mlp-scatter-test}
\end{figure}

\begin{figure}[htb]
\centering
\includegraphics[width=\linewidth, page=2]{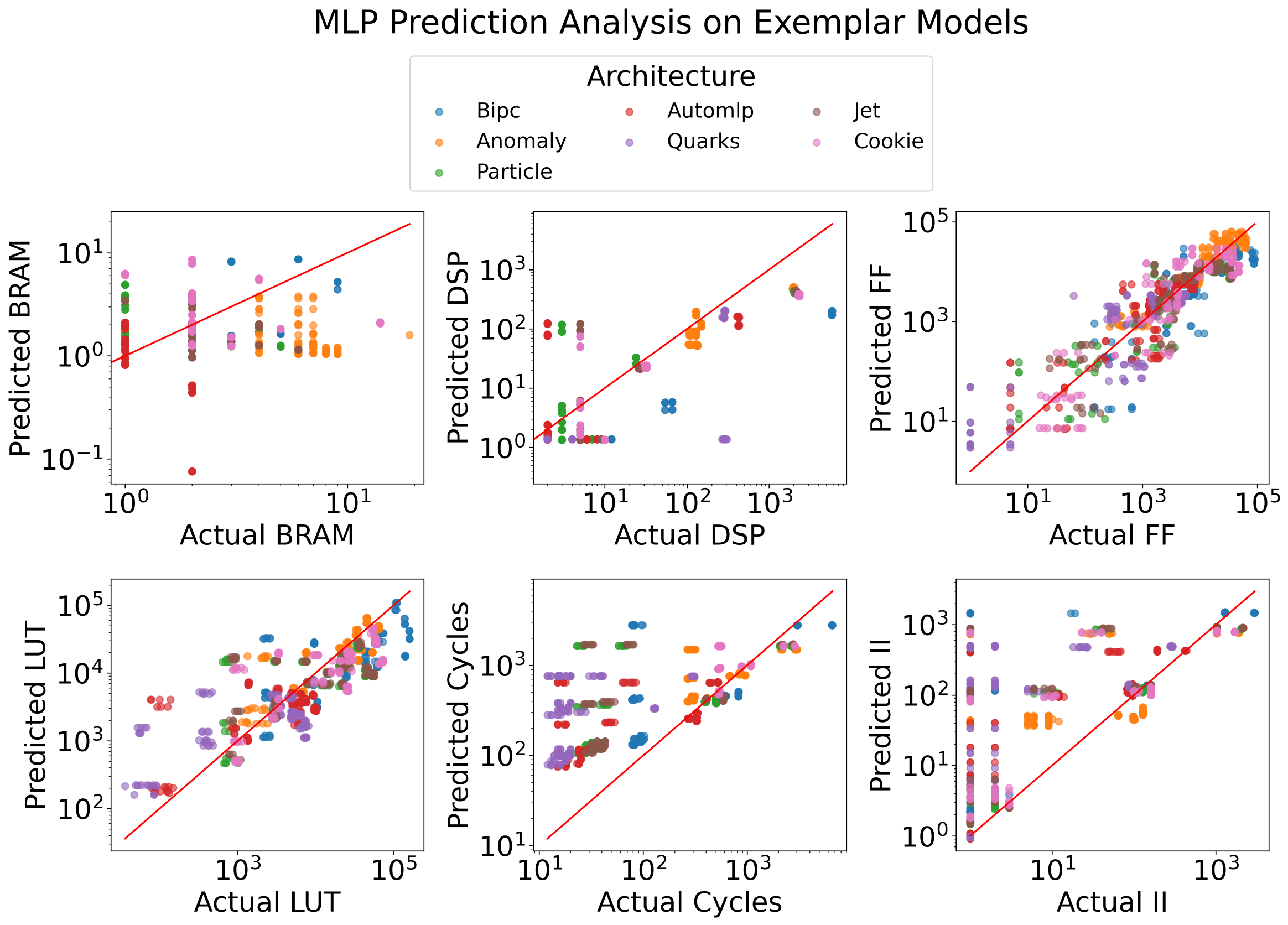}
\caption{Scatter plots of MLP predictions on the exemplar models. The red line shows the deviation from true values. Both axes are set to a logarithmic scale.}
\label{fig:mlp-scatter-benchmark}
\end{figure}

\begin{figure}[htb]
\centering
\includegraphics[width=\linewidth, page=1]{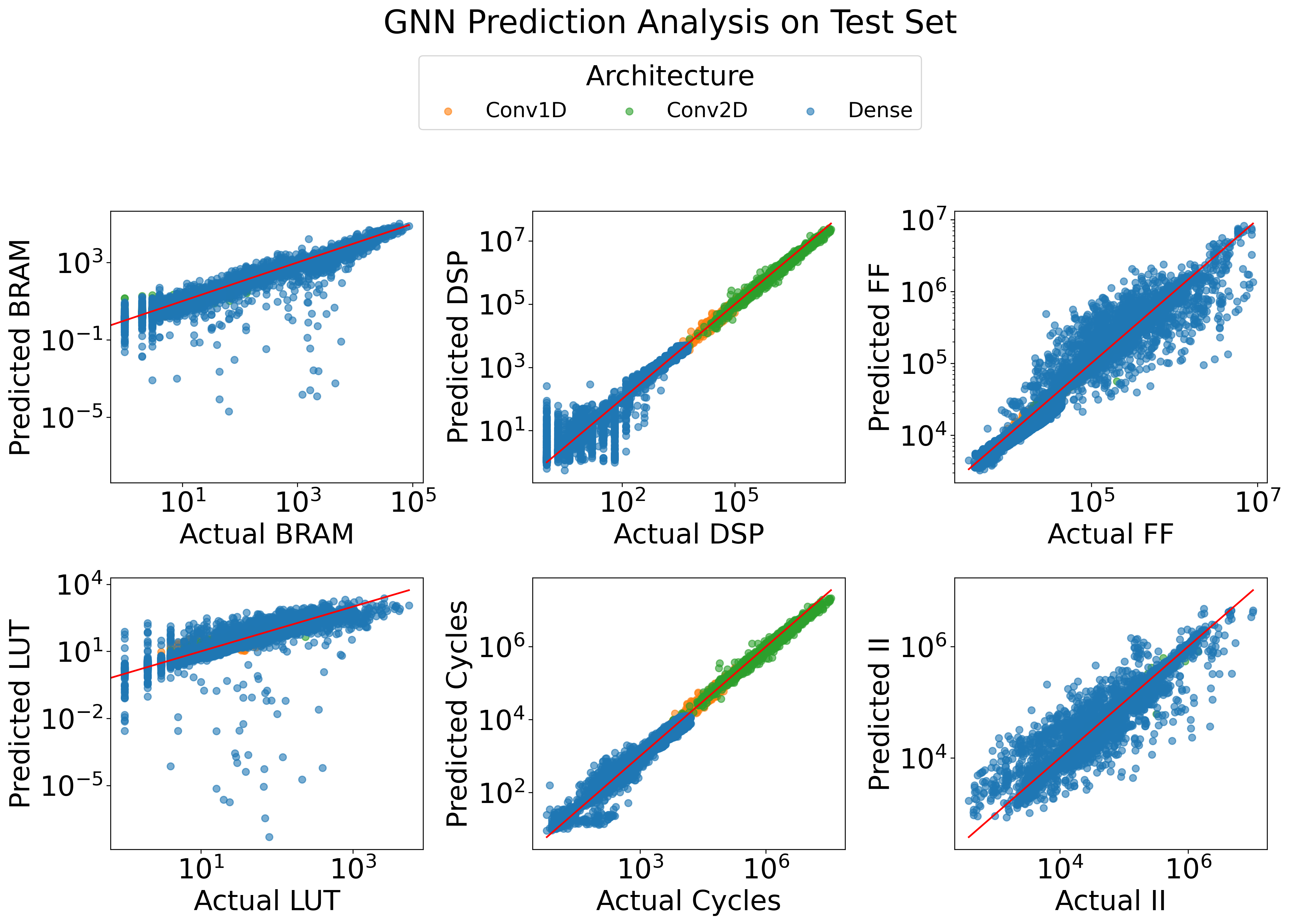}
\caption{Scatter plots of GNN predictions on the test set. The red line shows the deviation from true values. Both axes are set to a logarithmic scale.}
\label{fig:gnn-scatter-test}
\end{figure}

\begin{figure}[htb]
\centering
\includegraphics[width=\linewidth, page=2]{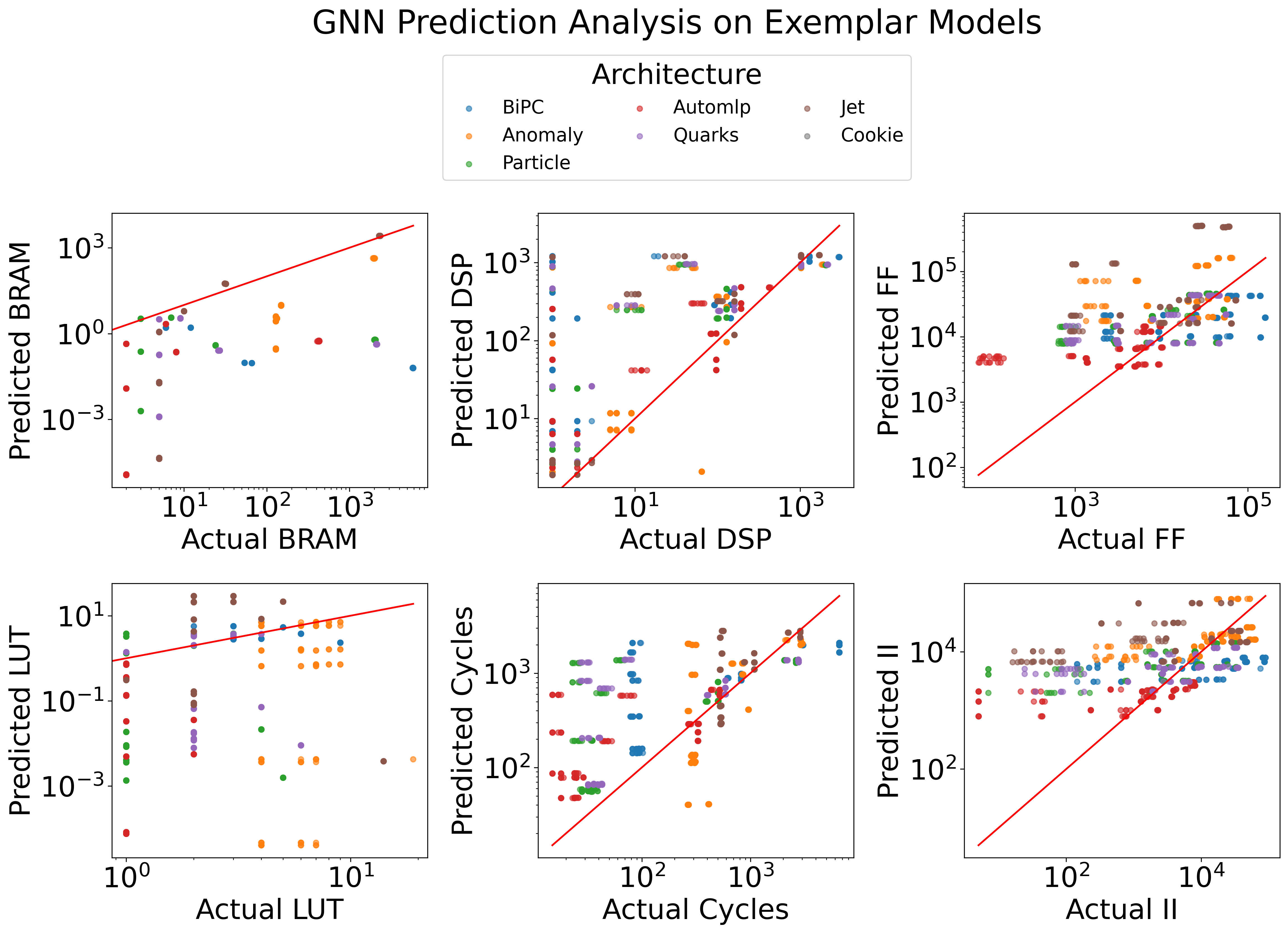}
\caption{Scatter plots of GNN predictions on the exemplar models. The red line shows the deviation from true values. Both axes are set to a logarithmic scale.}
\label{fig:gnn-scatter-benchmark}
\end{figure}

\begin{figure}[htb]
\centering
\includegraphics[width=\linewidth, page=1]{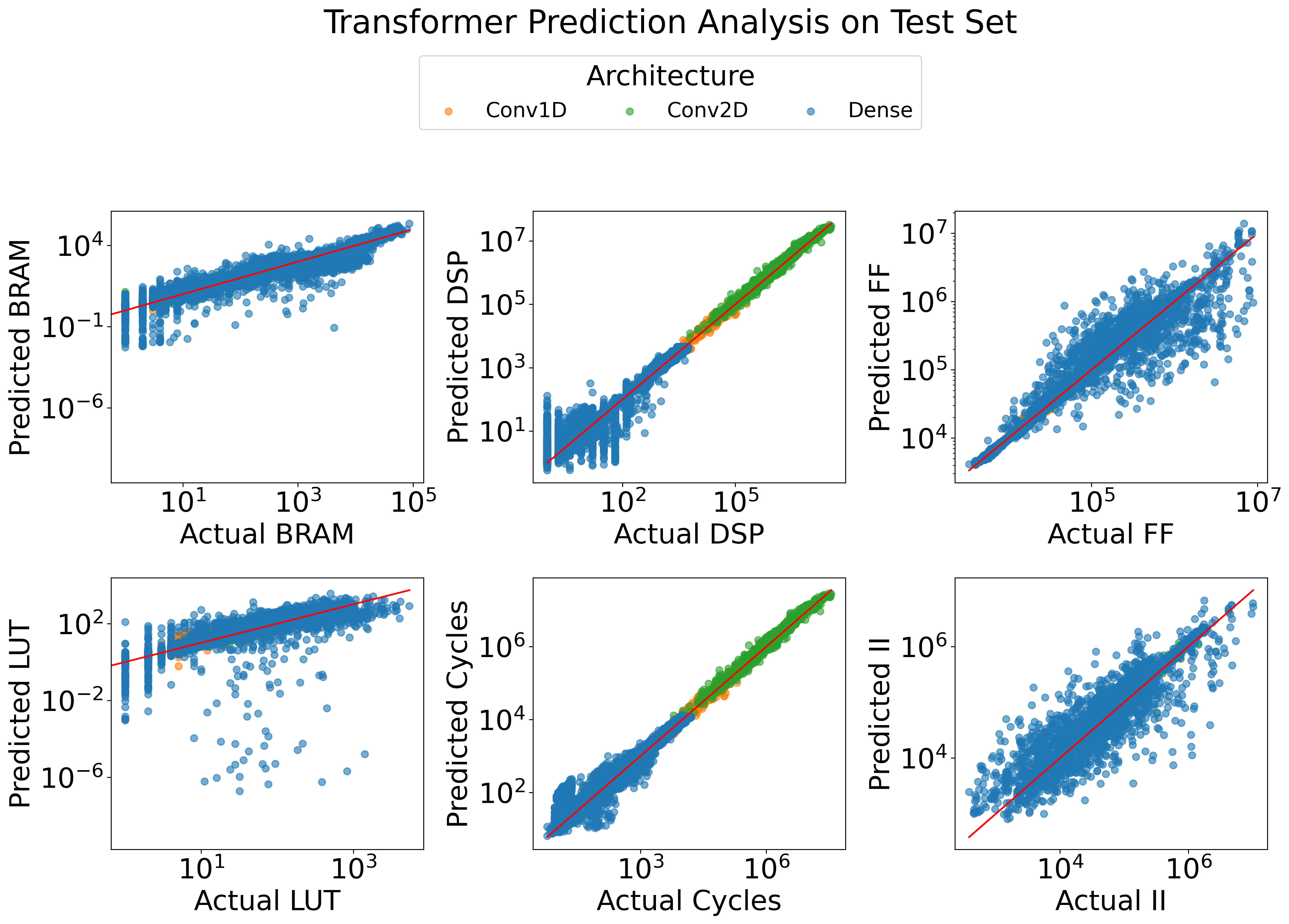}
\caption{Scatter plots of transformer predictions on the test set. The red line shows the deviation from true values. Both axes are set to a logarithmic scale.}
\label{fig:transformer-scatter-test}
\end{figure}

\begin{figure}[htb]
\centering
\includegraphics[width=\linewidth, page=2]{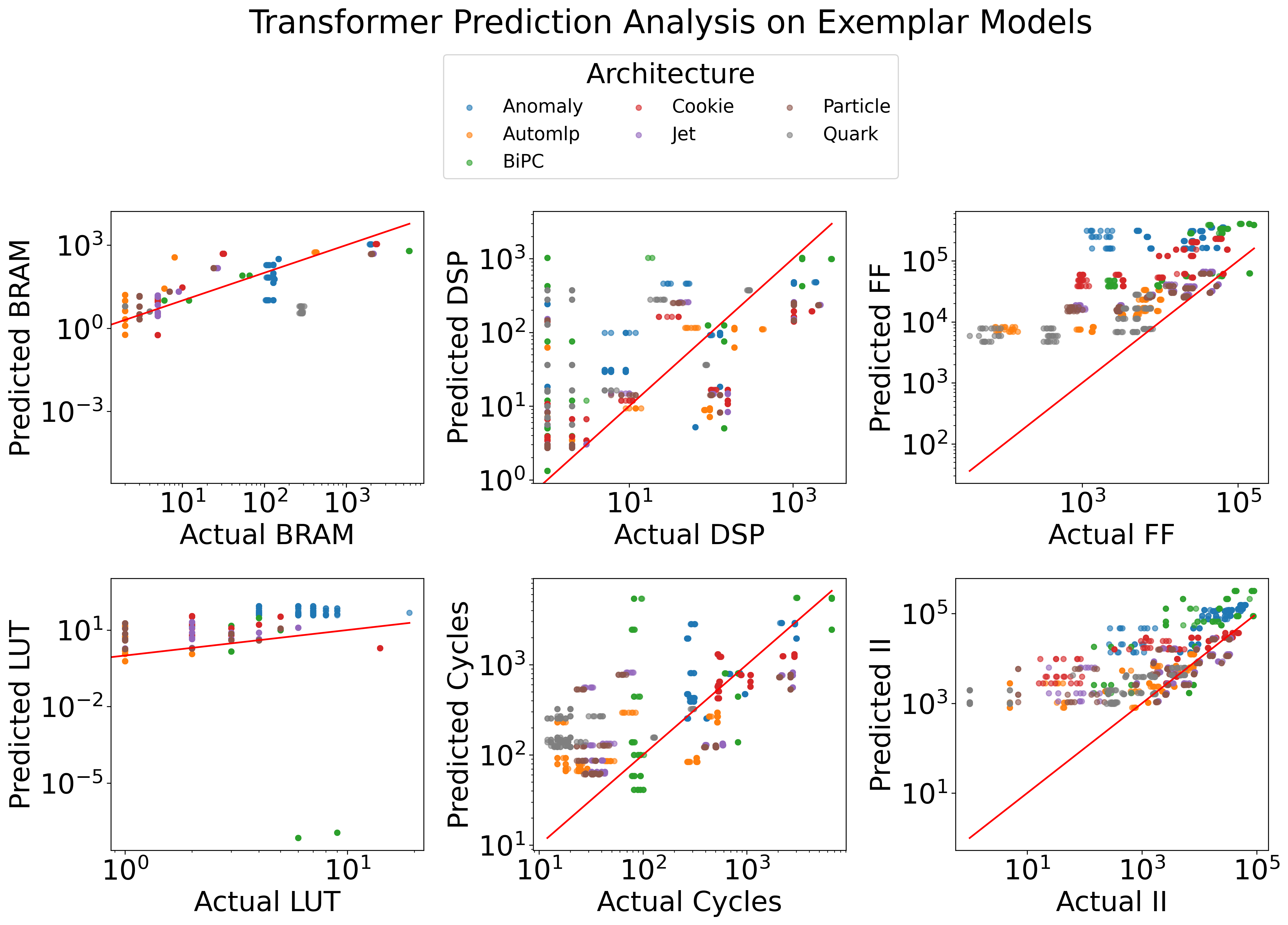}
\caption{Scatter plots of transformer predictions on the exemplar models. The red line shows the deviation from true values. Both axes are set to a logarithmic scale.}
\label{fig:transformer-scatter-benchmark}
\end{figure}
\end{appendices}

\end{document}